\documentclass[manuscript, screen]{acmart}

\usepackage{booktabs}

\usepackage[utf8]{inputenc}
\usepackage{amssymb}
\usepackage{amsmath}
\usepackage[boxruled,vlined,linesnumbered]{algorithm2e}

\usepackage{tabularx}
\usepackage{mathtools}
\usepackage{graphicx}
\usepackage{graphicx}
\usepackage[FIGTOPCAP]{subfigure}
\usepackage{enumitem}
\usepackage[font=small,labelfont=bf]{caption}
\usepackage{color}

\DeclarePairedDelimiter\ceil{\lceil}{\rceil}

\graphicspath{/figs}
\begin{document}

\title{Hybrid Isolation Forest - Application to Intrusion Detection}

\author{Pierre-Fran\c{c}cois Marteau}
\affiliation{%
  \institution{IRISA, CNRS, Universite Bretagne Sud}
  \city{Vannes}
  \country{France}}
\author{Saeid Soheily-Khah}
\affiliation{%
  \institution{IRISA, CNRS, Universite Bretagne Sud}
  \country{France}}
\author{Nicolas Béchet}
\affiliation{%
  \institution{IRISA, CNRS, Universite Bretagne Sud}
  \country{France}}

\begin{abstract}
From the identification of a drawback in the Isolation Forest (IF) algorithm  that limits its use in the scope of anomaly detection, we propose two extensions that  allow to firstly overcome the previously mention limitation and secondly to provide it with some supervised learning capability. The resulting Hybrid Isolation Forest (HIF) that we propose is first evaluated on a synthetic dataset to analyze the effect of the new meta-parameters that are introduced  and verify that the addressed limitation of the IF algorithm is effectively overcame. We hen compare the two algorithms on the ISCX benchmark dataset, in the context of a network intrusion detection application. Our experiments show that HIF outperforms IF, but also challenges the 1-class and 2-classes SVM baselines with computational efficiency. 
\end{abstract}


%\thanks{This work is supported by the National Science Foundation,  under grant CNS-0435060, grant CCR-0325197 and grant EN-CS-0329609.
%}

\maketitle

\section{Introduction}
Anomaly detection has been a hot topic for several decades and has led to numerous applications in a wide range of domains, such as fault tolerance in industry, crisis detection in finance and economy, health diagnosis, extreme phenomena in earth science and meteorology, atypical celestial object detection in astronomy or astrophysics, system intrusion in cyber-security, etc.

Anomaly detection is generally defined as the problem of identifying patterns that deviates from a  'normality' behavioral model.  In the literature, most approaches can be categorized either according to the model of normality that is involved or to the way they address the abnormality characterization and its identification. A quite exhaustive, although a bit old,  review in anomaly detection has been proposed in \cite{Chandola:2009}, completed by a more recent comparative study \cite{Goldstein2016}. According to these studies, the state of the art methods can be distributed into five main categories:

\begin{enumerate}
\item \textbf{Near neighbors and clustering based methods} \cite{Mennatallah2012}:  Near Neighbors methods rely on the assumption that a 'normal' instance occurs close to its near neighbors while an anomaly occurs far from its near neighbors. Similarly, cluster based methods rely on the assumption that a 'normal' instance occurs near its closest cluster centroid while an anomaly will occur far from its nearest cluster centroid \cite{key:articleBama,Lin201513}. 
\item \textbf{Classification based method}: in this paradigm, several classes of 'normal' data are learned by a set of one against all classifiers (each classifier is associated to a class and is trained to separate it from the others classes). An instance that is not categorized as 'normal' by any of these classifiers is considered as an anomaly. A peculiar case occurs when a single class is used to model the 'normal' data.  Random Forest, including recent advances on one-class random forest \cite{Desir2013}, multi-class and one-class Support Vector Machine (SVM) \cite{Fujimaki2008}, and neural networks \cite{Lee:1998:DMA:1267549.1267555,Ghosh:1999:SUN:1251421.1251433,ryan:nips10}, are the most used classifiers for anomaly detection. 
\item \textbf{Statistical based methods} rely on the assumption that 'normal' data are associated to high probability states of an underlying stochastic process while anomalies are associated to low probability states of this process. Popular approaches in this category are kernel based density models and  the  Gaussian Mixture Model (GMM), including recent advance in one-class GMM \cite{Kemmler2013}, 
\item \textbf{Information theoretic based methods} use  information theoretic measures \cite{Lee2001}, such as the entropy, the Kolmogorof complexity, the minimum description length, etc,  to estimate the 'complexity' of the 'normal' dataset (or equivalently the complexity of the process behind the production of these data) \cite{GFDLS2006}. If $\mathcal{C}(D)$ estimates the complexity of dataset $D$, the minimal subset of instance $I$ that maximizes $\mathcal{C}(D) - \mathcal{C}(I)$ is considered as the anomaly subset.
\item \textbf{Spectral based method} rely on the assumption that it is possible to embed the data into a lower dimensional subspace in which 'normal' data and anomalies are supposedly well separated \cite{Akoglu2015}. PCA and graph (of similarity) clustering are among the most popular methods in this category.
\end{enumerate} 

A recent review \cite{Agrawal2015}, though much less detailed, identifies a supplemental category that stands in between supervised and unsupervised methods, and that authors refer to as 'hybrid methods'. Such approaches combine supervised, semi-supervised and unsupervised techniques.

In 2008, Isolation Forest (IF) \cite{Liu2008}, a quite conceptually different approach to the previously referenced methods, has been proposed that went strangely below the radar of the previous review. The IF paradigm is based on the \textit{difficulty} to isolate a particular instance inside the whole set of instances when using (random) tree structures. It relies on the assumption that an anomaly is in general much easier to isolate than a 'normal' data instance. Hence, IF is an unsupervised approach that relates somehow to the information theoretic based methods since the \textit{isolation difficulty} is addressed through an algorithmic complexity scheme. IF has been successfully used in some applications, \cite{Liu2012}, \cite{Ding2013} and extended recently in \cite{Shen2016} to improve the selection of attributes and their split values when constructing the tree.  The main advantage of IF based algorithms is their capability to process large amount of data in high dimension spaces with computational and spatial controlled efficiency compared to other unsupervised methods. Unfortunately, as we shall see in the next section, IF suffers from what we call "blind spots", namely empty portions of the space in which data instances are embedded, that are nevertheless considered as normality spots by the IF algorithm.

Apart from identifying this drawback, the aim of this paper is to propose, firstly, a semi-supervised extension of the IF algorithm to overcome the 'blind spot' limitation, and secondly a supervised extension in order to benefit from known anomalies if any. The Hybrid Isolation Forest (HIF) is thus a hybrid method that takes advantage of the unsupervised paradigm of IF while correcting a weakness that could have very damaging effects, and furthermore benefits from a supervised learning capability to enhance the effectiveness of the anomaly detection. 

We detail the IF algorithm in the second section of this paper, and give some highlights about the occurrence of the so-called 'blind spots' by using a synthetic dataset. The third section presents the extension of the IF algorithm that we propose and shows, on the previous synthetic dataset, how this extension can be used to get rid of blind spots. The supervised functionality that we add to the IF is also described by the end of this section. The fourth section addresses an application in the domain of intrusion detection that assesses in a close to real-life situation the benefits brought by the HIF algorithm. Our results show that the proposed HIF algorithm compares advantageously with the state of the art baselines in anomaly detection that we have considered, namely one-class and two-classes SVM.

\section{Isolation forest and its 'blind spot'}

The simple idea behind the isolation forest approach, is that it is (in general) much simpler to isolate an 'outlier' from the rest of the data than to isolate an 'inlier' from the rest of the data.

\begin{figure}[h!]
    \centering
    \begin{tabular}{ccc}
        \centering
        \includegraphics[scale=0.30]{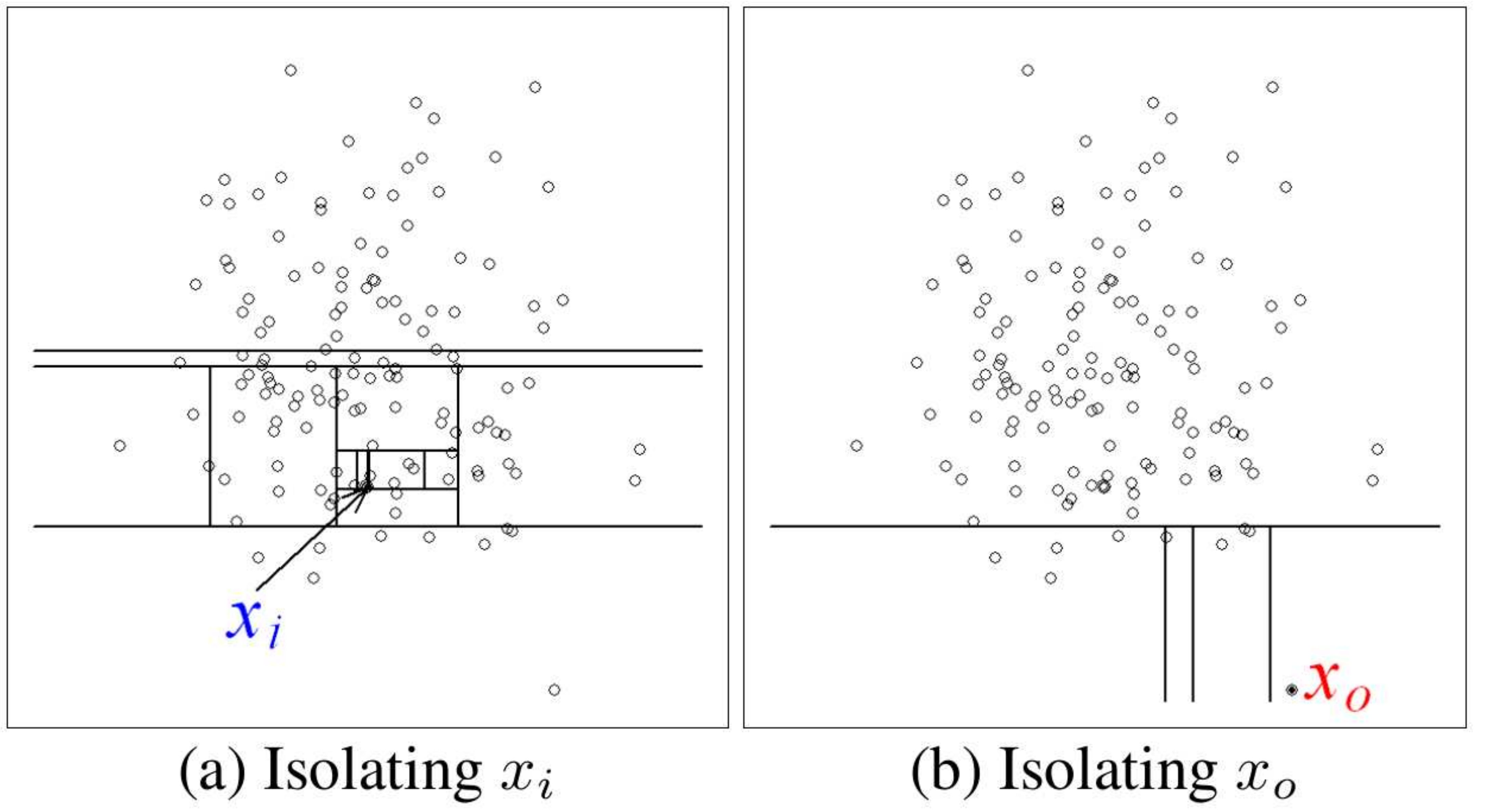}
       \end{tabular}
        \caption{Principle of the IF algorithm (Figure is from \cite{Liu2008}). $x_i$ is an 'inlier', while $x_o$ is an 'outlier' (anomaly).}
        \label{fig:IF_principle}
    \end{figure}

This leads, in the context of a binary tree partitioning algorithm, to expect a shorter path to locate an 'outlier' and a longer path to locate an 'inlier'. This is exemplified in Fig.\ref{fig:IF_principle}, which shows that, for a 2D normally distributed dataset, more separating lines are needed to separate the 'inlier' $x_i$ from the rest of the data comparatively to the number of separating lines needed to isolate the 'outlier' $x_0$.

\subsection{The Isolation Forest algorithm}

We reproduce hereinafter the description of the isolation forest algorithm as presented in \cite{Liu2008}.\\

\subsubsection{Building the isolation forest:}
Let $X \subset \mathbb{R}^d  $ be the set of instances. 
The IF algorithm is an ensemble based approach that builds a forest of random binary trees. Given a sample $S$ randomly drawn from $X$, an isolation tree $iT(S)$ is recursively built according to the (iTree) algorithm \ref{Algo:iTree}:

\begin{algorithm}[]
\DontPrintSemicolon
\SetAlgoLined
%\KwResult{an iTree}
\SetKwInOut{Input}{Input}\SetKwInOut{Output}{Output}
\Input{$S \subset X$, $l$ the current depth level, $l_{max}$ the maximal depth limit}
\Output{an iTree}
\BlankLine
 
\If{$l \ge l_{max}$ or $|S| \le 1$}
    {
    \Return exNode($S$)\;
    }
    \Else{
    randomly select a dimension $q \in \{1, \cdots, n\}$\;
    randomly select a split value $p$ between max and min
	values along dimension $q$ in $S$\;
	$S_l \leftarrow $ filter$(S, q < p)$\;
	$S_r \leftarrow $ filter$(S, q \ge p)$\;
	\Return inNode(Left $\leftarrow$ iTree$(S_l , l + 1, l_{max})$,\;
	\hspace{10mm} Right $\leftarrow$ iTree$(S_r , l + 1, l_{max})$,\;
	\hspace{10mm} splitDim $\leftarrow q$,\;
	\hspace{10mm} splitVal $\leftarrow p$)\;
   } 		 
\caption{Function iTree$(S, l, l_{max})$}
\label{Algo:iTree}
\end{algorithm}

Two meta-parameters are required to build an isolation forest: $\psi$, the size of the subsets $S$ that are used to build the trees, and $t$, the number of trees. Parameter $l_{max}$, the maximum height of the trees, is empirically set up to $\ceil{log_2 \psi}$. 

Finally, the isolation forest $iF=\{iT(S_1), iT(S_2), \cdots, iT(S_t)\}$ is obtained by randomly selecting $\{S_1, S_2, \cdots, S_t\}$, $t$ samples in $X$ with $|S_i|=\psi$ for all $i$, and constructing an iTree on each of these samples, as depicted in Algorithm \ref{Algo:iTree}.\\

\subsubsection{Constructing the anomaly score: }

The anomaly score for the isolation forest is constructed from the analysis of unsuccessful search in a Binary Search Tree (BST). 
The path length $h(x)$ for data point $x$ is defined as
the number of edges $x$ that traverses an iTree from the root node
until the traversal is terminated at an external node.
For a data set of $n$ instances, Section 10.3.3 of \cite{Preiss1999} gives the average path length of unsuccessful search in BST as:

\begin{equation}
c(n) = 2H(n - 1) - (2(n - 1)/n)
\label{eq:c(n)}
\end{equation}

where $H(i)$ is the harmonic number and that can be estimated 
by $log(i) + 0.5772156649$ (Euler’s constant). As $c(n)$ is the
average of $h(x)$ given $n$, we use it to normalize $h(x)$. The
anomaly score $s$ of an instance $x$ is defined as:
\begin{equation}
s(x, n) = 2^{-\frac{E(h(x))}{c(n)}}
\label{eq:ifScore}
\end{equation}

where $E(h(x))$ is the mean of $h(x)$ taken over a collection of
isolation trees. One can easily establish that:
\begin{itemize}
\item when $E(h(x)) \rightarrow c(n)$, $s \rightarrow 0.5$;
\item  when $E(h(x)) \rightarrow 0$, $s \rightarrow 1$;
\item  and when $E(h(x)) \rightarrow n - 1$, with $n$ large, $s \rightarrow 0$.
\end{itemize}

Finally, once the isolation forest is constructed, given any $x \in \mathbb{R}^d$, an anomaly score for $x$, $s(x)$, is calculated according to Eq. \ref{eq:ifScore}, where $E(h(x))$ is evaluated on the set of iTrees of the forest.\\

The Algorithm \ref{Algo:pathLength} presents the recursive evaluation of $h(x)$ given $x$ and an iTree, $e$, the current path length, being initialized with $0$.\\

\begin{algorithm}[]
\DontPrintSemicolon
\SetAlgoLined
%\KwResult{a path length}
\SetKwInOut{Input}{Input}\SetKwInOut{Output}{Output}
\Input{$x$ an instance, $T$ an iTree, $e$ the current path length; to be initialized to zero when first called}
\Output{h(x), path length for $x$}
\BlankLine

\If{$T$ is an external node}
    {
    \Return $e + c(T.size)$ ($c(.)$ is defined in Equation (\ref{eq:c(n)}))\;
    }
$a \leftarrow T.splitAtt$\;
\If{$x[a] < T.splitValue$}
	{
	\Return PathLength$(x, T.left, e + 1)$\;
	}
	\Else{ 
	%\Comment {x[a] ≥ T.splitValue}\;
	\Return PathLength$(x, T.right, e + 1)$\;
	}
\caption{Function PathLength$(x, T, e)$}
\label{Algo:pathLength}
\end{algorithm}

\subsection{Existence of 'blind spots' in IF based anomaly detection}
The assumption behind the IF algorithm is that anomalies will be associated to short paths in the iTrees, leading to a high anomaly score $s$, while 'normal' data will be associated to longer paths, leading to a low anomaly score $s$. Unfortunately, if this is true for normally distributed data for instance, this is not true in general. In particular, this assumption is not verified for data distributed in a concave set such as a tore or a set with a 'horse shoe' shape. To demonstrate this, we develop the following test based on synthetic data.
   
\subsubsection{Synthetic experiment} 
\begin{figure}[h!]
    \centering
    \subfigure[]{
        \includegraphics[scale=0.24]{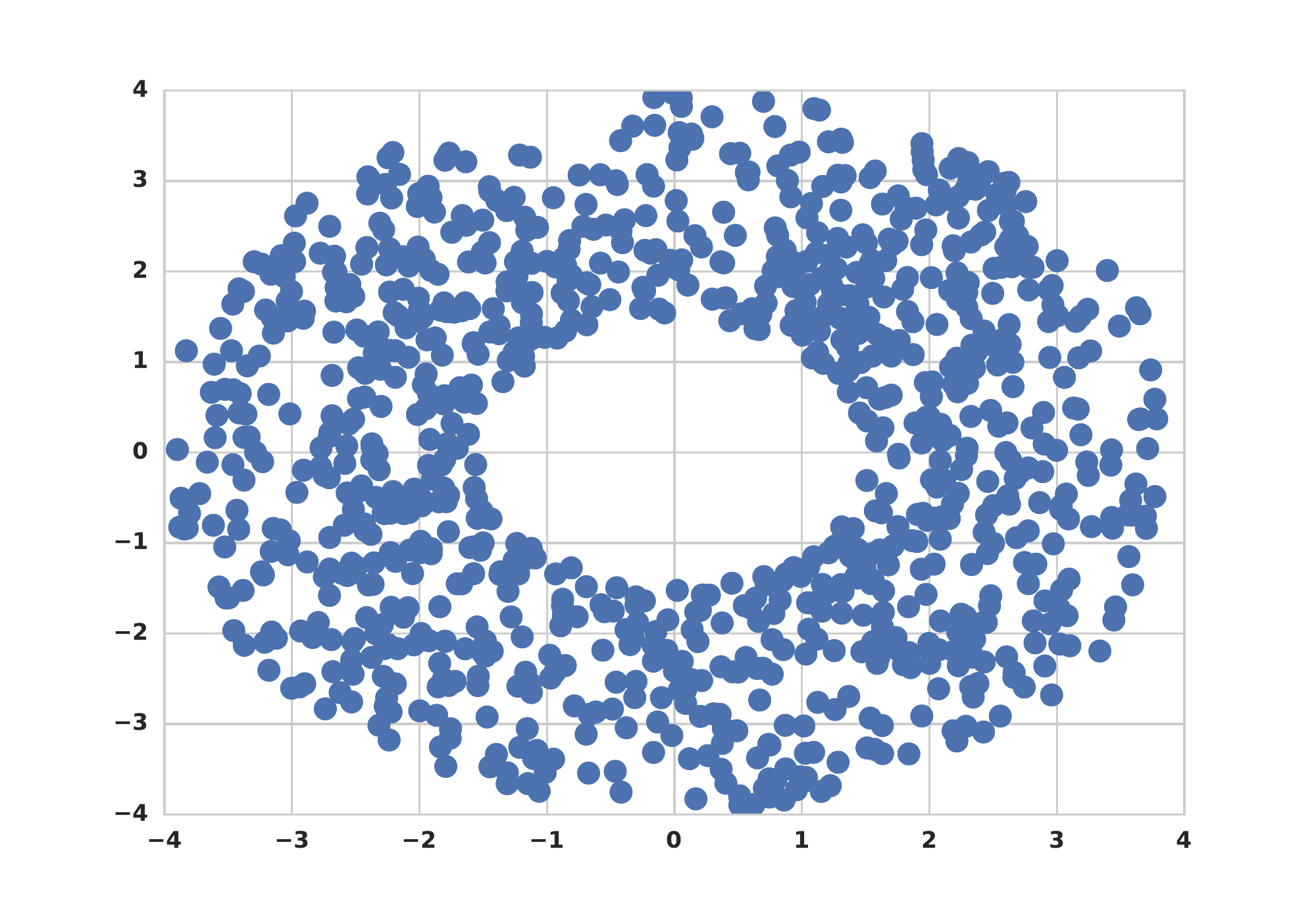} 
    }
    ~
     \subfigure[]{
        \includegraphics[scale=0.24]{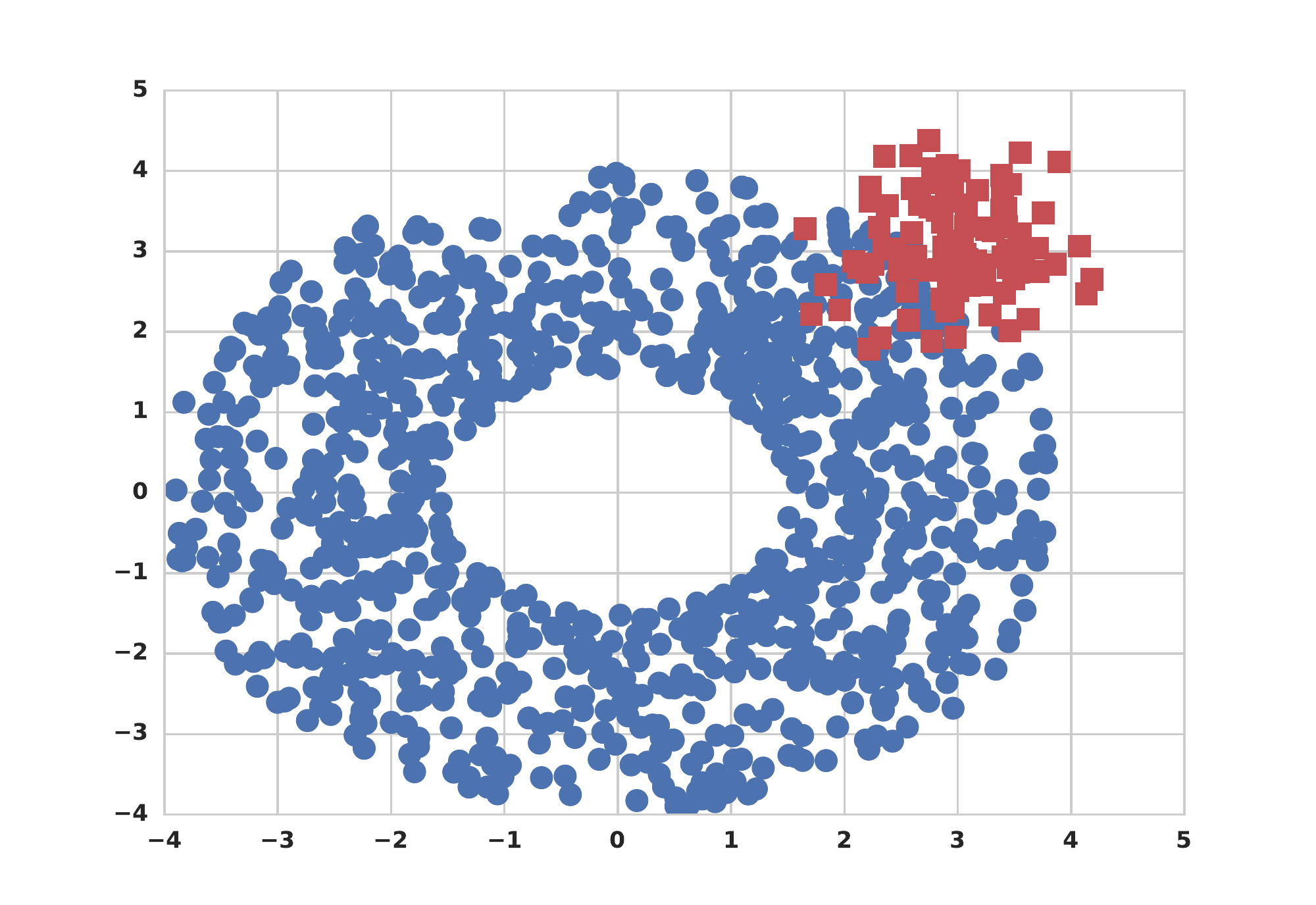} 
    }
    ~
     \subfigure[]{
        \includegraphics[scale=0.24]{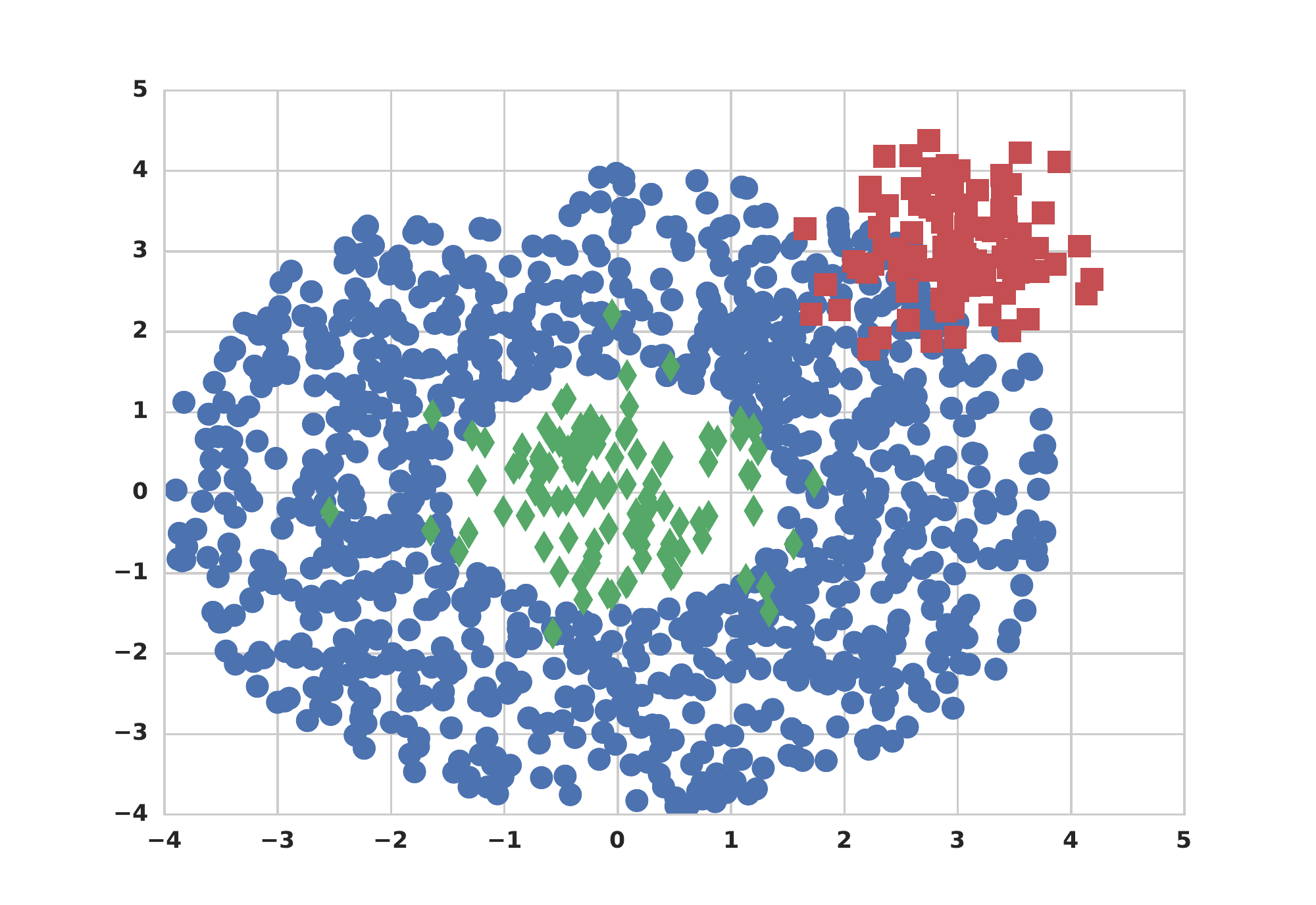} 
    }

  \caption{2D-tore 'normal' data set in blue round dots (a), with a cluster of anomaly data in red square dots at the top right side of the tore (b), with an additional cluster of anomaly data in green diamond dots at the center of the tore (c).}
  \label{fig:clusters}
 \end{figure}

For this setting, 'normal' data belongs to a 2D-tore centered in $(0,0)$ and delimited by two concentric circles whose radius are respectively $1.5$ and $4$. A training ($X_n$) and a testing ($X_{nt}$) sets of normal data are uniformly drawn from this 2D-tore, each one containing $n=1000$ instances, as depicted in Fig. \ref{fig:clusters} (a). 

A first 'anomaly' set ($X_{r}$) is drawn from a Normal distribution with mean $(3., 3.)$ and covariance $((.25, 0), (0, .25))$, as depicted in red square dots in \ref{fig:clusters} (b). These anomalies intersect the 2D-tore at its top right side.

A second 'anomaly' set ($X_{g}$) is drawn from a Normal distribution with mean $(0., 0.)$ and covariance $((.5, 0), (0, .5))$, as depicted in green diamond dots in \ref{fig:clusters} (c). These anomalies are located at the center of the 2D-tore.
 
\begin{figure}[h!]
    \centering
    \subfigure[]{
        \includegraphics[width=0.5\textwidth]{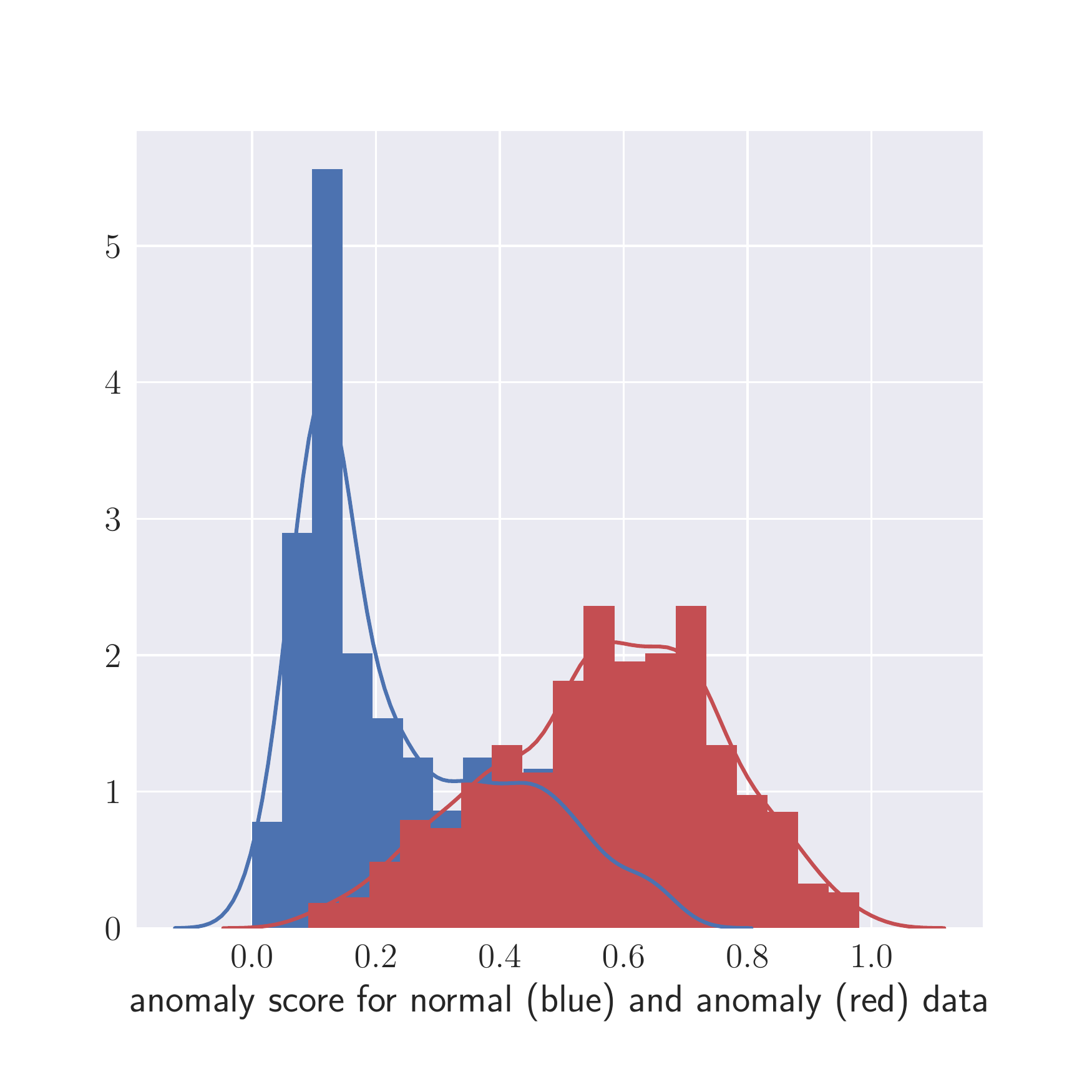} 
            }
    ~
     \subfigure[]{
        \includegraphics[width=0.5\textwidth]{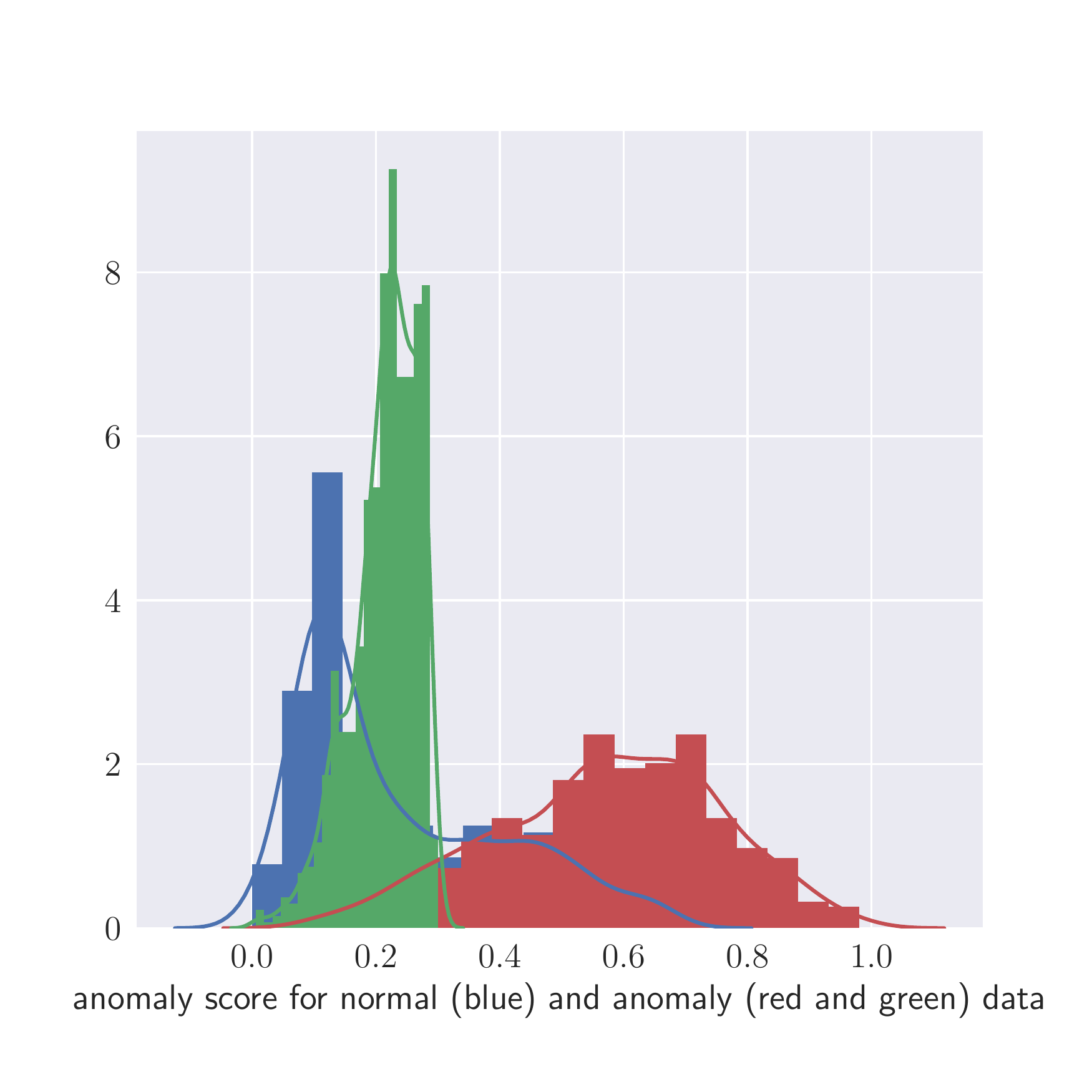}
            }
    \caption{Distributions of the anomaly scores for the 'normal' data, in blue, and for the red anomalies (a), and with the insertion of the green anomalies (b).}
 \label{fig:distrib1}
\end{figure}

Then we build the IF (with $\psi=64$ and $t=512$) from the 'normal' dataset $X_{n}$ and evaluate the distributions of the anomaly scores obtained for the 'normal' 'blue' test dataset $X_{nt}$, the 'red' anomalies $X_{r}$ and the 'green' anomalies $X_{g}$. Fig.\ref{fig:distrib1} presents the 'normal' v.s. 'red' anomalies (a)  distributions, and with the addition of the green anomaly distribution (b). 

At this point, we clearly show that the 'green' anomaly distribution is in large intersection with the 'normal' data distribution, which is not the case for the 'red' anomaly distribution. Hence anomalies located at the center of the tore are likely to be much more mis-detected by the IF algorithm than anomalies located at the periphery of the tore.\\

 \begin{figure}[h!]
    \centering
     \subfigure[]{
        \includegraphics[width=0.5\textwidth]{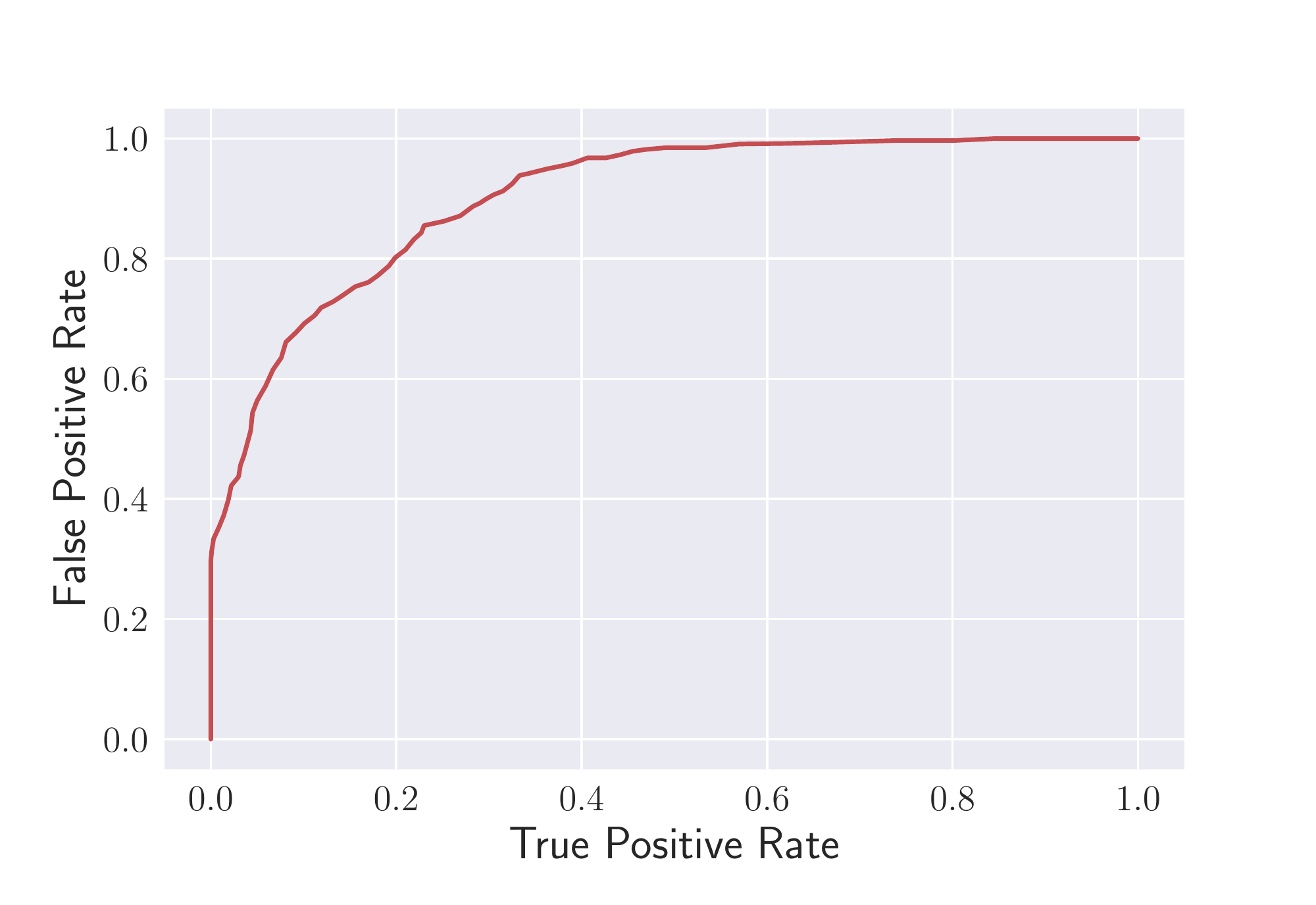} 
            }
    ~
     \subfigure[]{
        \includegraphics[width=0.5\textwidth]{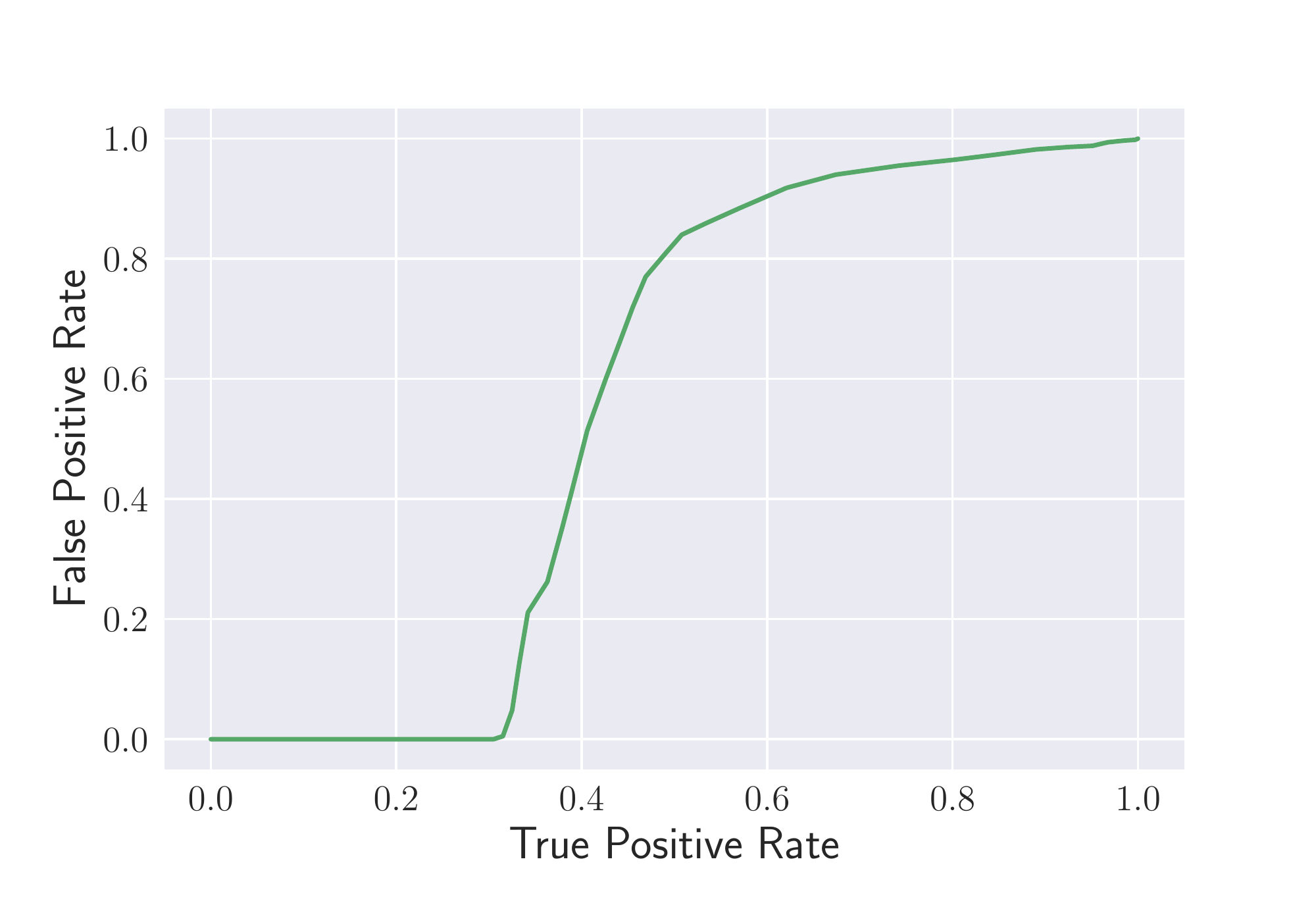} 
    }
    \caption{ROC curves for anomaly detectors constructed using the IF scores, 'normal' test data against the red anomalies (a), and against the green anomalies (b).}
    \label{fig:roc1}
\end{figure}

Fig.\ref{fig:roc1} presents the Receiver Operating Curve (ROC) for an anomaly detector based on the scores provided by an IF trained on 'normal' data only. On this test, the Area Under the Curve measures (AUC) are respectively  $0.925$ and $0.441$ for the 'red' and 'green' anomalies respectively. Hence, the IF does not perform better than a random classifier when it comes to separate 'green' anomalies from 'normal' data. This is what we call a 'blind spot' effect. \\

To fix this mis-detection problem, several approaches can be carried out. The first one consists in finding a new embedding of the data in which 'blind spots' are not present. For the 2D-tore experiment, one can easily replace the 2D Cartesian coordinate system by a 2D polar system and get rid of the 'blind spot'. However, such explicit embedding does not exists or is difficult to elicit in general. In the same line of approaches,  Kernel based methods define an implicit embedding that allows for projecting the data in a higher dimensional space (possibly an infinite dimensional one) in which the hope is that the tackled problem becomes (more) linearly separable. The kernel PCA \cite{Scholkopf1998} is a good example of such method. The difficulty is that the implicit embedding is defined through  the so-called kernel gram matrix, that can be very hard to compute in the context of large volume of data.  

As 'blind spots' are obviously associated to some correlation into the variables that describe the data, the second avenue to avoid them is to find a transformation so that the new variables are statistically independent, or as independent as possible. Unfortunately, if linear correlation is easily removable, nonlinear correlation is much more difficult to remove. This can be feasible to some extend and in some framework, such as blind source separation, though it requires some knowledge about the data, such as the number of sources \cite{Jutten1991}. For problems expressed in high dimension it is often possible to reduce the dimension using an appropriate embedding that allows for describing the data with variables that are weakly correlated, but in general at the cost of losing information (the embedding is not invertible) [REFs] \cite{Bunte201223,Carreira-perpinan_theelastic,Zheng2014,DBLP:journals/tgrs/LungaE13} .

\section{Hybrid isolation forest }

Complementary to the previously mentioned approaches, we propose to overcome the 'blind spot' drawback of IF by adding, directly into the IF framework itself, two simple sources of information that will allow for designing a more sophisticated anomaly score:
\begin{enumerate}
\item the first one is an unsupervised extension that exploits a distance knowledge to neighboring 'normal' data, 
\item the second one is a supervised-based extension that exploits a distance knowledge to neighboring 'anomaly' data. 
\end{enumerate} 

\subsection{Adding a distance-based score}

In the 2D-tore experiment, the 'green' anomalies are undetected because they are located inside an area that is as difficult to 'isolate' using a BST than 'normal' data. However, most of these data are located at a greater distance to the 'normal' data associated to the external node bucket found by the BST algorithm when evaluating the path length (see line 2 of Algorithm \ref{Algo:pathLength}).

Hence, the simple idea we propose to overcome the 'blind spot' effect consists in integrating this distance based information into the IF score as follows: 
\begin{itemize}
\item first the centroid of the data associated to each external node of the hiTree is evaluated when building the hiTree (training phase). This corresponds to lines 2-5 in Algorithm \ref{Algo:hiTree}.
\item for each tested data, its distance to the centroid of the external node found by the BST algorithm is evaluated and reported as a second score element.  This corresponds to line 3 in Algorithm \ref{Algo:hiScore}.
\end{itemize}  

The final score $s_c(x)$ provided at the forest level is simply the expectation of the $\delta(x)$ scores evaluated on each of the iTrees of the forest.

\begin{equation}
s_c(x) = E(\delta(x)))
\label{eq:anomalyScore}
\end{equation}

\begin{algorithm}[]
\DontPrintSemicolon
\SetAlgoLined
%\KwResult{an iTree}
\SetKwInOut{Input}{Input}\SetKwInOut{Output}{Output}
\Input{$S \subset X$, $l$ the current depth level, $l_{max}$ the maximal depth limit}
\Output{an hiTree}
\BlankLine
 
\If{$l \ge l_{max}$ or $|S| \le 1$}
    {
    $C_S$=None\;
    \If{$|S| \ge 0$}{
    	$C_S$=Centroid($S$)\;
    }
    \Return exNode($S$, $SLab$)\;
    }
    \Else{
    randomly select a dimension $q \in \{1, \cdots, n\}$\;
    randomly select a split value $p$ between max and min
	values along dimension $q$ in $S$\;
	$S_l \leftarrow $ filter$(S, q < p)$\;
	$S_r \leftarrow $ filter$(S, q \ge p)$\;
	\Return inNode(Left $\leftarrow$ iTree$(S_l, SLab_l, l + 1, l_{max})$,\;
	\hspace{10mm} Right $\leftarrow$ iTree$(S_r, SLab_r, l + 1, {max})$,\;
	\hspace{10mm} splitDim $\leftarrow q$,\;
	\hspace{10mm} splitVal $\leftarrow p$)\;
   } 		 
\caption{Function hiTree$(S, SLab, l, l_{max})$}
\label{Algo:hiTree}
\end{algorithm}

\begin{algorithm}[]
\DontPrintSemicolon
\SetAlgoLined
%\KwResult{an iTree}
\SetKwInOut{Input}{Input}\SetKwInOut{Output}{Output}
\Input{$x \in \mathbb{R}^d$: an anomaly, $x_{lab}$: a label attached to the anomaly, $T$: a node in an iTree}
\Output{none}
\BlankLine
 
\If{$T$ is an external node}
    {
	append $x_{lab}$ to the list of anomaly labels attached to T ($T.labels$) \;
	append $x$ to list of anomalies attached to ($T.X_{a}$) \;
    }
    \Else{
	$a \leftarrow T.splitAtt$\;
	\If{$x[a] < T.splitValue$}    	
		{
    	return addAnomaly($x$, $x_{lab}$, $T.left$)
    	}
    	\Else{
    	return addAnomaly($x$, $x_{lab}$, $T.right$)
    	}
   } 		 
\caption{Function addAnomaly($x$, $x_{lab}$, $T$)}
\label{Algo:addAnomaly}
\end{algorithm}

\subsection{Adding known anomalies}

In some situations, some anomalies may have been detected and documented. This is the case for any monitored environment in which incident are reported such as in network supervision and intrusion detection for instance. In such situations, incorporating this kind of expert knowledge into the IF framework can be a source for improvements for the detection of new anomalies. Furthermore, this could offer a way to better balance, according to the targeted application, the anomaly mis-detection rate  and the false alarm rate. 

To implement this supervised extension to the isolation forest algorithm, at the training phase, we develop a function, 
detailed in algorithm \ref{Algo:addAnomaly}, that assigns a known (labeled) anomaly $x_a$ to the external nodes (the leaf of the iTree) that is returned when searching the BST for $x_a$. By the end of this process, each external node stores in a list of all its assigned anomalies and associated labels (line 2-3 of the algorithm). 

Once all the anomalies have been introduced in the iTrees, to complete the training, for each external node we evaluate the centroid of the anomalies that have been attached to it, as depicted in algorithm \ref{Algo:computeAnomalyCentroid}.

At the testing phase, we finally evaluate a dedicated score $\delta_a(x)$ for each iTree as shown in line 4 of the scoring algorithm (Algorithm \ref{Algo:hiScore}).  Given the external node $T$ of an iTree found when searching for the tested instance $x$, $\delta_a(x)$ corresponds to the distance between $x$ and the centroid of the set of anomalies assigned to $T$ ($T.X_c$).

The final supervised score contribution $s_a(x)$ provided by the HIF is simply the ratio of the expectation of  $\delta(x)$ (the distance of $x$ to the local centroid of 'normal' data, Eq. \ref{eq:anomalyScore}) over the expectation of $\delta_a(x)$, basically the ratio of the mean of the scores $\delta_a(x)$ over the mean of the scores $\delta_a(x)$ evaluated on each of the iTrees of the forest as stated in Eq. \ref{eq:scorea}.  If $E(\delta_a(x))=0$, then we simply enforce $s_a(x)=0$.

\begin{equation}
s_a(x)=\frac{E(\delta(x))}{E(\delta_a(x))}=\frac{Mean_{iTree}\delta(x, iTree)}{Mean_{iTree}\delta_a(x, iTree)} 
\label{eq:scorea}
\end{equation}

\begin{algorithm}[]
\DontPrintSemicolon
\SetAlgoLined
%\KwResult{an iTree}
\SetKwInOut{Input}{Input}\SetKwInOut{Output}{Output}
\Input{$T$: a node in an iTree}
\Output{none}
\BlankLine
 
\If{$T$ is an external node}
    {
    $C_a$=None \;
   \If{$|T.X_a| \ge 0$}{
    	$C_a$=Centroid($T.X_a$)\;
    	}
    }
    \Else{
		computeAnomalyCentroid($T.left$)\;
		computeAnomalyCentroid($T.right$)\;
    	}		 
\caption{Function computeAnomalyCentroid($T$)}
\label{Algo:computeAnomalyCentroid}
\end{algorithm}

%    def updateCa(self):
%        if self.ntype == 'exNode':
%            if(len(self.Xanomaly)>0):
%                self.Ca = np.mean(self.Xanomaly, axis=0)
%        else:
%            self.left.updateCa()
%            self.right.updateCa()

\begin{algorithm}[]
\DontPrintSemicolon
\SetAlgoLined
%\KwResult{a path length}
\SetKwInOut{Input}{Input}\SetKwInOut{Output}{Output}
\Input{$x$ an instance, $T$ an hiTree, $e$ the current path length; to be initialized to zero when first called}
\Output{$h(x)$, the path length for $x$, $\delta(x)$, the Euclidean distance between $x$ and the centroid associated to the external node, and $\delta_a(x)$ the minimal Euclidean distance between $x$ and the centroid of the anomalies attached to $T$, $T.X_a$}
\BlankLine

\If{$T$ is an external node}
    {
    $h(x) = e + c(T.size)$ ($c(.)$  is defined in Equation  (\ref{eq:c(n)}))\;
    $\delta(x)$ = EuclideanDistance$(x, T.C_S)$\;
    $\delta_a(x)$ = EuclideanDistance$(x,T.C_a)$\;
    \Return $h(x)$, $\delta(x)$, $\delta_a(x)$\;
    }
$a \leftarrow T.splitAtt$\;
\If{$x[a] < T.splitValue$}
	{
	\Return hiScore$(x, T.left, e + 1)$\;
	}
	\Else{ 
	%\Comment {x[a] ≥ T.splitValue}\;
	\Return hiScore$(x, T.right, e + 1)$\;
	}
\caption{Function hiScore$(x, T, e)$}
\label{Algo:hiScore}
\end{algorithm}

\subsection{Aggregating the scores}

The IF score $s(x,n)$ for data $x$ and dataset size $n$, given in Eq.(\ref{eq:ifScore}) need to be aggregated with the two new scores $s_c(x)$ and $s_a(x)$. We adopt a simple aggregation in two steps:
\begin{enumerate}
\item first, all the three scores are normalized such as to fit into the unit interval $[0;1]$. On the train data we evaluate for $s(x) \in \{s(x,n), s_c(x), s_a(x) \}$ 
\begin{equation}
\tilde{s}(x) = \frac{s(x)-min(s(x))}{max(s(x))-min(s(x))}
\label{eq:normalization}
\end{equation}
\item then we introduce two meta-parameters $\alpha_1$ and $\alpha_2$ $\in [0;1]$ and evaluate the following linear model:
\begin{equation}
s_{hif}(x,n) = \alpha_2 \cdot \Big(\alpha_1 \cdot \tilde{s}(x,n) + (1 - \alpha_1) \cdot \tilde{s_c}(x)\Big) + (1 - \alpha_2) \cdot \tilde{s_a}(x))
\label{eq:aggregation}
\end{equation}
We discuss the selection or optimization of these meta parameters in section \ref{sec:meta_alpha1_alpha2}.
\end{enumerate}

\subsection{Back to the synthetic experiment}
We go back to the synthetic experiment to assess the impact of the new scoring involved in the HIF algorithm.

\begin{figure}[!h]
    \centering
    \begin{tabular}{ccc}
        \centering
        \includegraphics[scale=0.3]{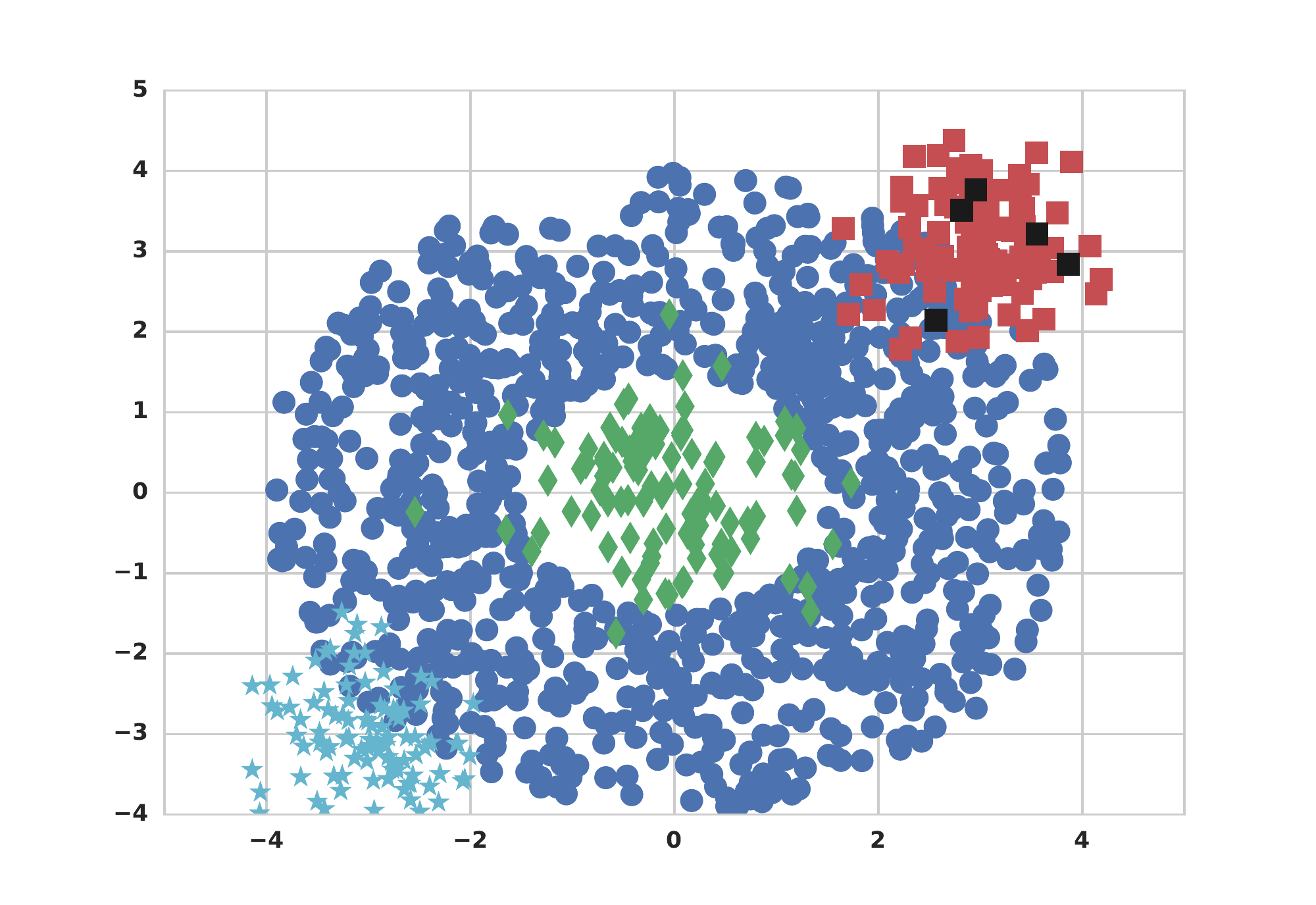}
        \end{tabular}
        \caption{2D-tore 'normal' dataset (blue), with three anomaly clusters in red dots (top right of the tore), green (center of the tore) and cyan dots (bottom left of the tore). To train the semi-supervised isolation forest, 0.5\% of the red anomalies are used and shown in black dots.}
        \label{fig:clusters2}
    \end{figure}
    
To the previous setting (Fig.\ref{fig:clusters}) we add a third anomaly cluster constructed from the Normal distribution with mean $(-3., -3.)$ and covariance $((.25, 0), (0, .25))$ ($X_c$, containing also $1000$ anomalies) in cyan star dots at the bottom left of the 2D-tore, as shown in Fig. \ref{fig:clusters2}. Some labeled 'red' anomalies are also represented in black dots in the figure. 

We construct the HIF with the same meta parameter setting that we previously used for the construction of the IF, namely $\psi=64$ and $t=512$.

\subsubsection{Impact of the $s_c(x)$ score}
\begin{figure}[!h]
    \centering
    \includegraphics[width=0.5\textwidth]{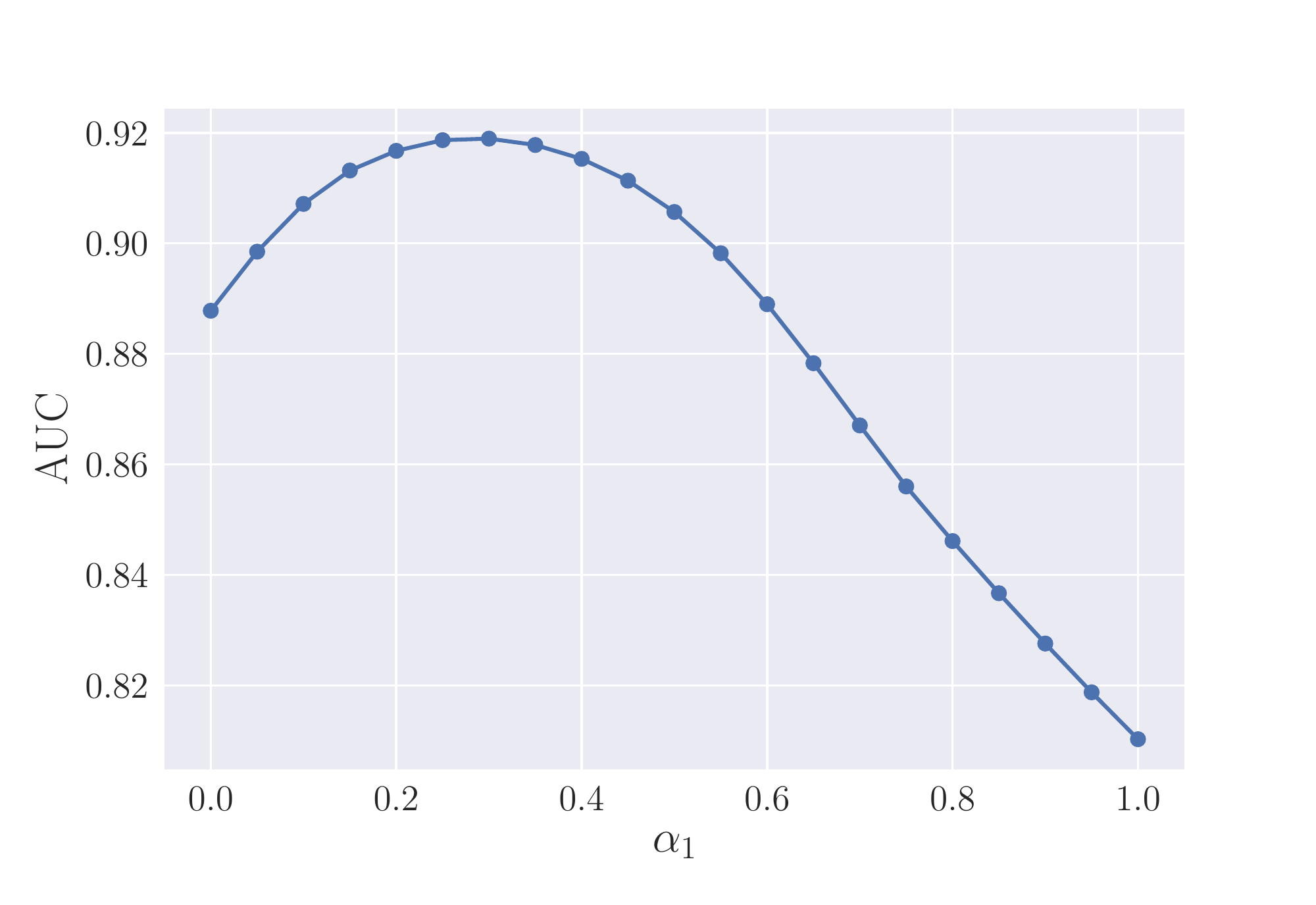} 
 \caption{AUC values as a function of $\alpha_1$ when the HIF score is evaluated according to Eq. \ref{eq:reduced_aggregation} (HIF1), and all the anomalies are considered in a single test. The maximal AUC value is obtained near $\alpha_1=.3$}
 \label{fig:auc_all_alpha1}
\end{figure}

: we consider at this stage that $\alpha_2=1$ in the HIF score, which leads to the reduced aggregation score given in Eq. \ref{eq:reduced_aggregation}. We refer to this configuration as the HIF1 algorithm. Hence none anomaly are added to the HIF yet. Figure \ref{fig:auc_all_alpha1} presents the AUC value as a function of $\alpha_1$ when all the test data are considered ('normal' test data are mixed up with all kind of anomalies). An optimum AUC value can be found near $\alpha_1=.3$.

 \begin{figure}[!ht]
    \centering
     \subfigure[]{
        \includegraphics[width=0.3\textwidth]{distribIF1.pdf} 
            }
    ~
     \subfigure[]{
        \includegraphics[width=0.3\textwidth]{distribIF11.pdf} 
    }
    ~
     \subfigure[]{
        \includegraphics[width=0.3\textwidth]{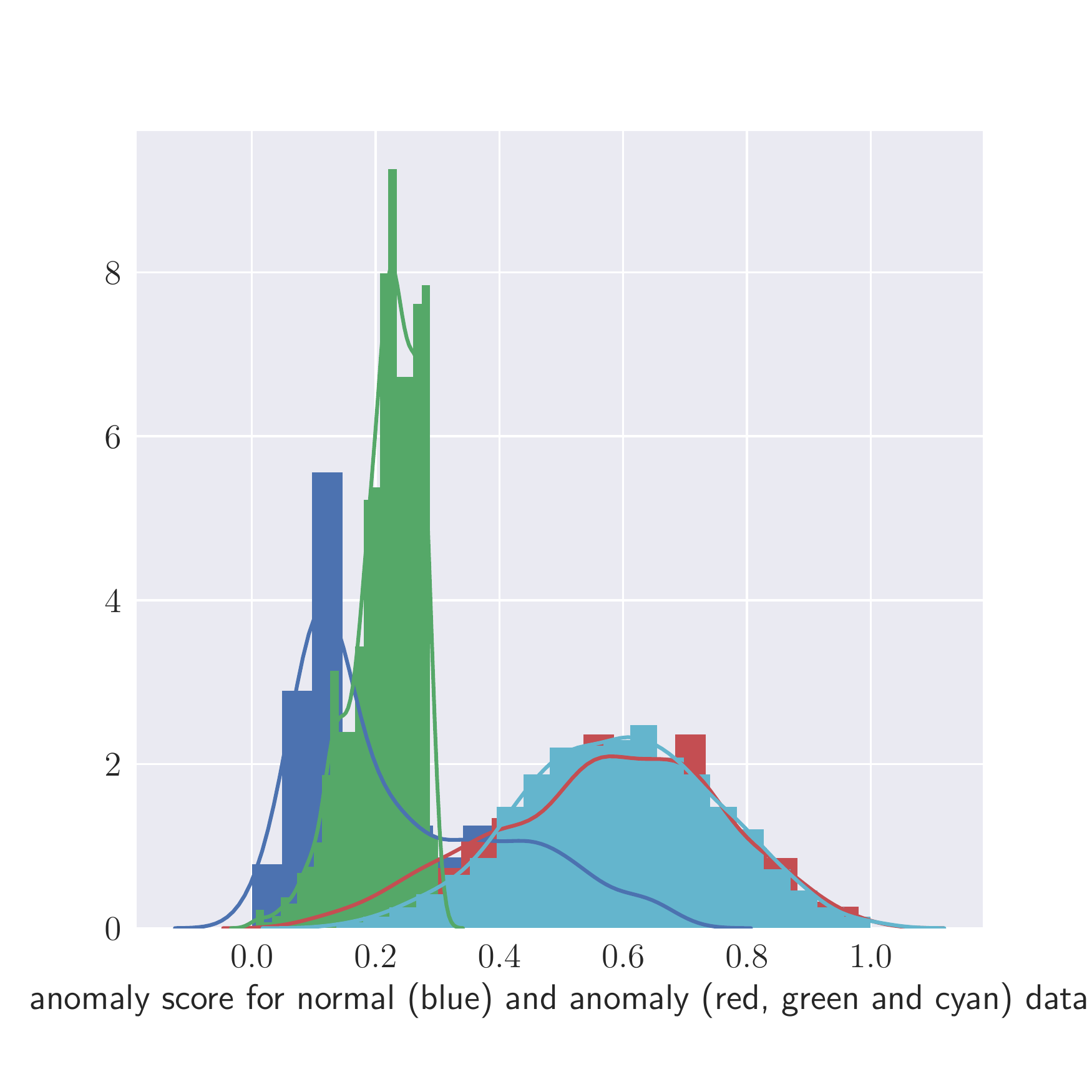} 
    }\\
    \subfigure[]{
        \includegraphics[width=0.3\textwidth]{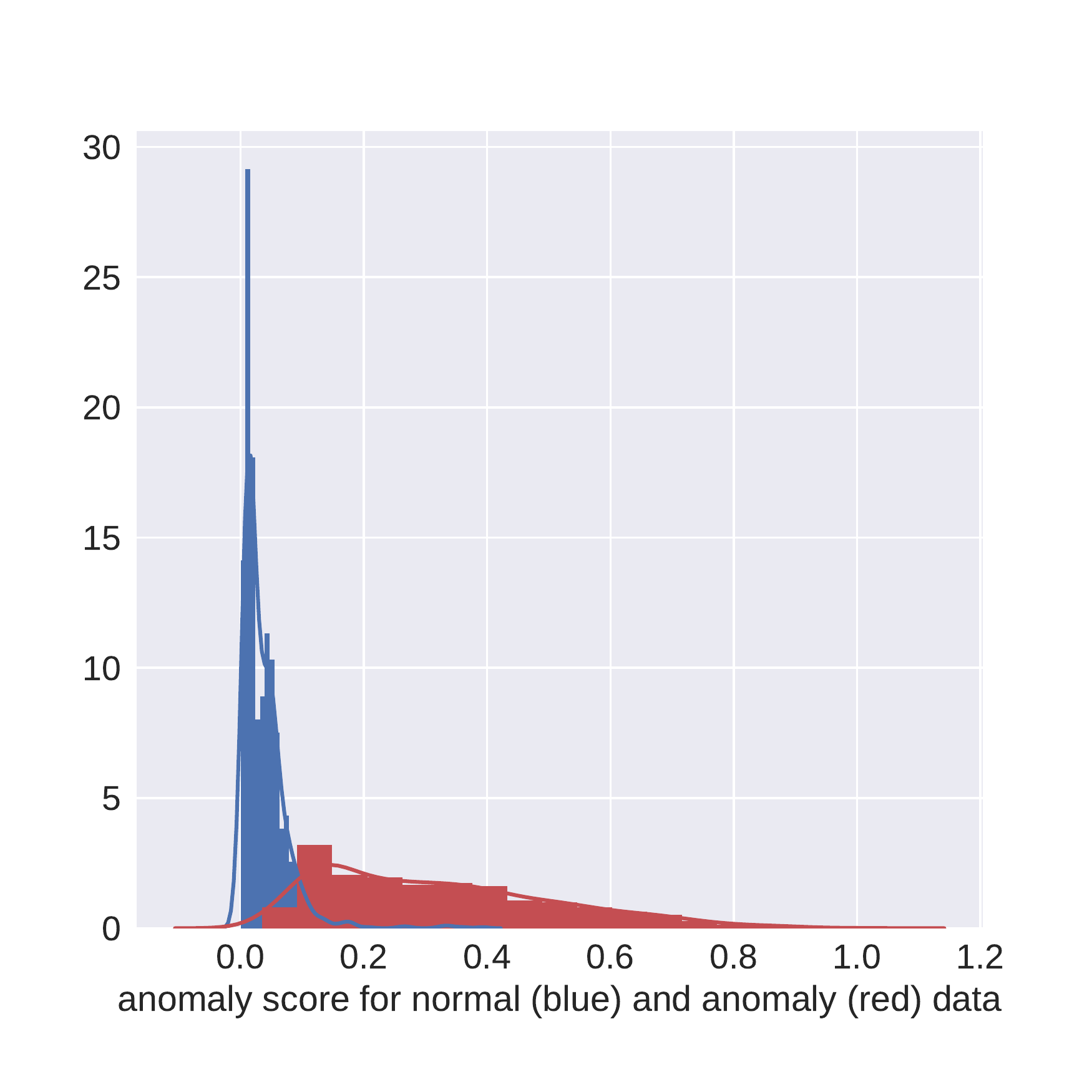} 
            }
    ~
     \subfigure[]{
        \includegraphics[width=0.3\textwidth]{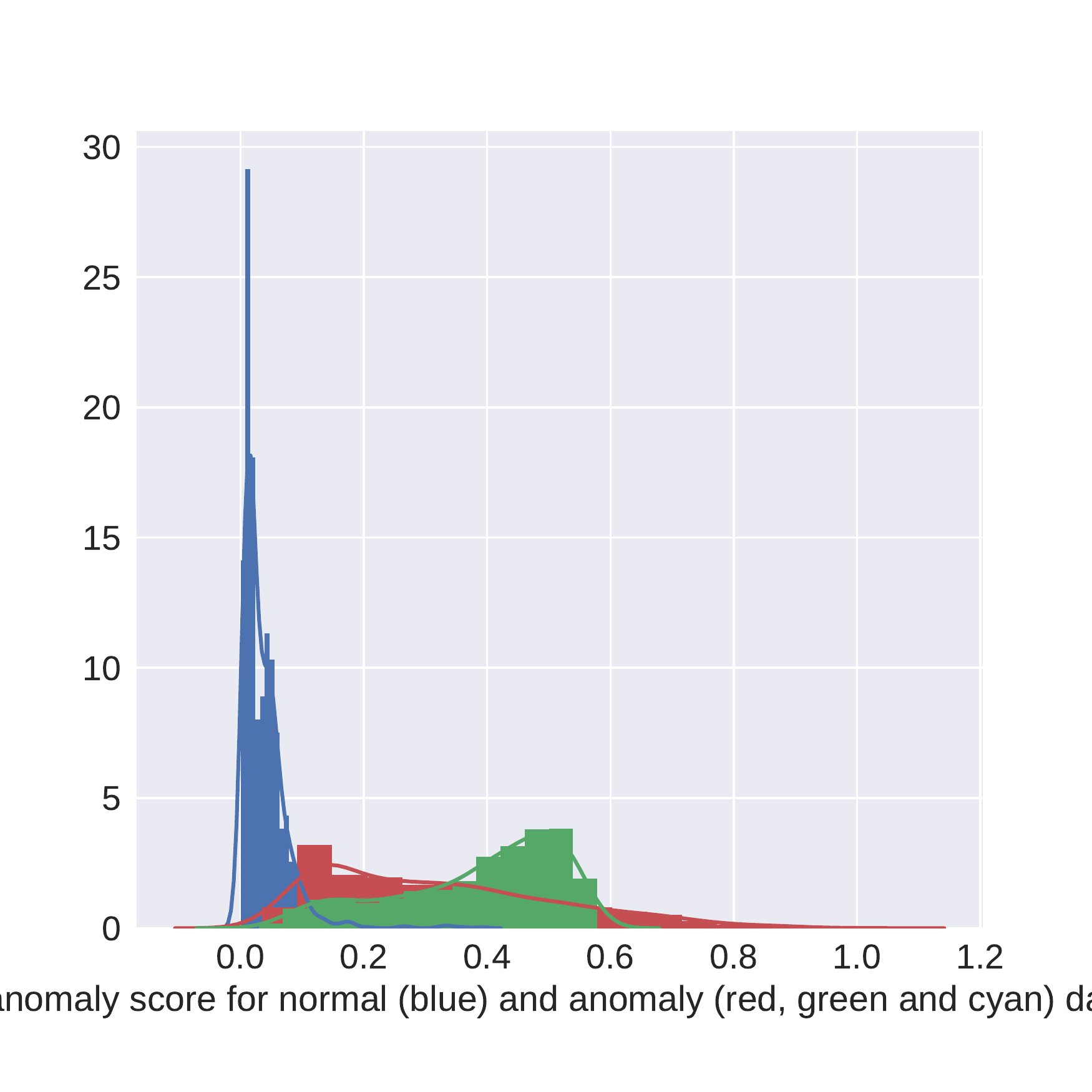} 
    }
    ~
     \subfigure[]{
        \includegraphics[width=0.3\textwidth]{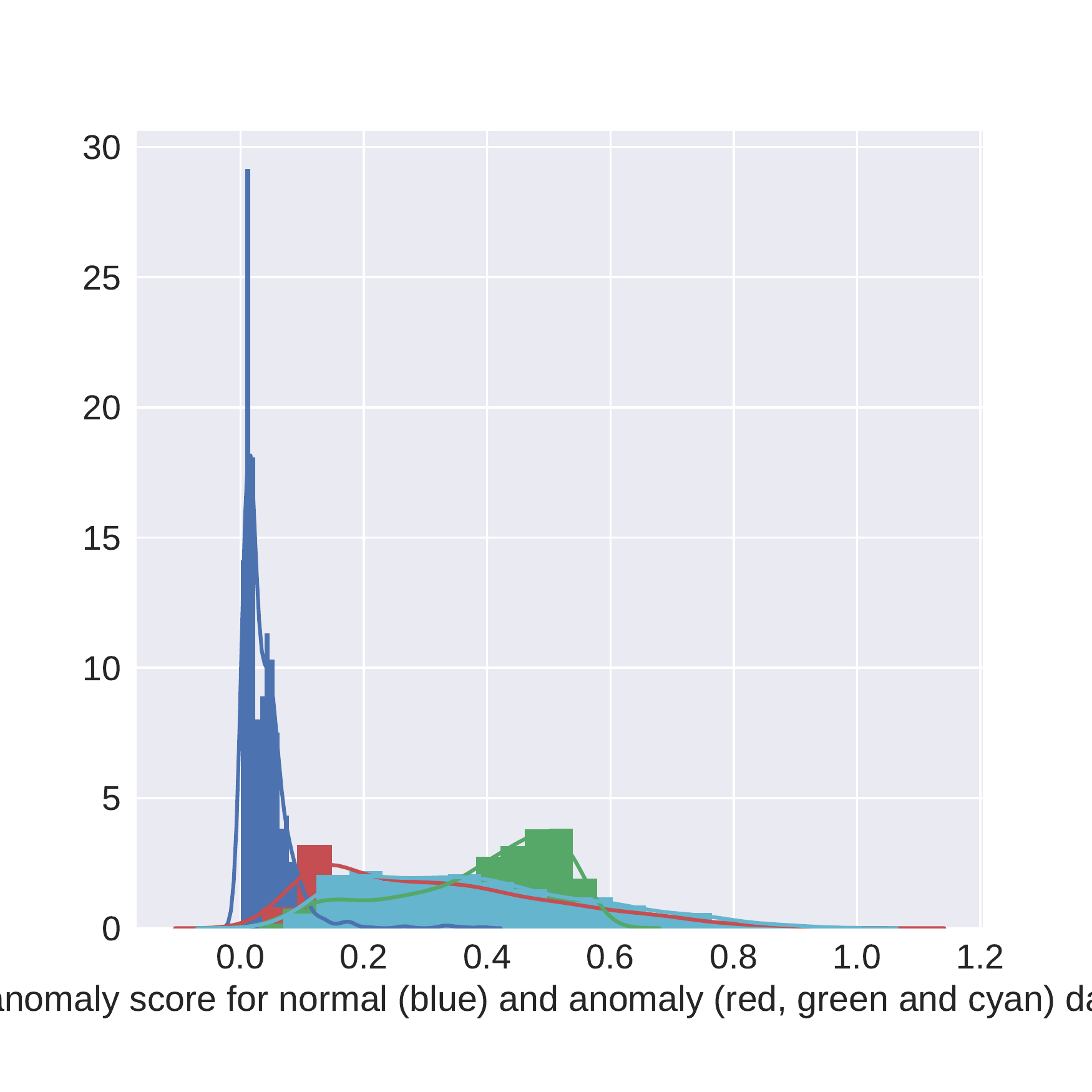} 
    }
    \caption{Distributions of the anomaly scores for the 'normal' data (in blue), and progressively superimposed for the red ((a), (d)), green ((b), (e)) and cyan ((c), (f)) anomaly clusters. Top: IF distributions, bottom: HIF1 distributions (none labeled anomalies have been added).}
    \label{fig:distrib2}
\end{figure}

\begin{equation}
s_{hif1}(x,n) = \alpha_1 \cdot \tilde{s}(x,n) + (1 - \alpha_1) \cdot \tilde{s_c}(x)
\label{eq:reduced_aggregation}
\end{equation}

Fig. \ref{fig:distrib2} presents the distributions of the scores obtained by the IF algorithm (top) and the HIF1 algorithm (bottom). Clearly, the 'green' anomaly cluster located at the center of the 2D-tore is much more separated for HIF1 than for IF. The 'red' and 'cyan' anomaly clusters distributions remain apparently well separated by the two algorithms from the 'normal' data distribution in blue. 

\begin{table}[h]
\centering
  \begin{tabular}{l|c||c|c|c|}
    &  IF &  HIF1 (Eq.\ref{eq:aggregation})& $\alpha_1$  \\
    \hline\hline
    Mean  &	.793 &	.937 &	.245	\\ \hline
    Std.Dev. & .007	& .009	& .009 	\\
    \hline
  \end{tabular}
  \caption{Mean value with standard deviations for best $\alpha_1$ ($\alpha2=1$), and corresponding mean AUC values with standard deviations for IF and HIF1 algorithms. For this experiment, 'normal' test data are mixed up with all the tested anomalies (red, green, cyan). }
 \label{tab:table_alpha1_ALL}
\end{table}

 \begin{figure}[h!]
    \centering
    \includegraphics[width=0.5\textwidth]{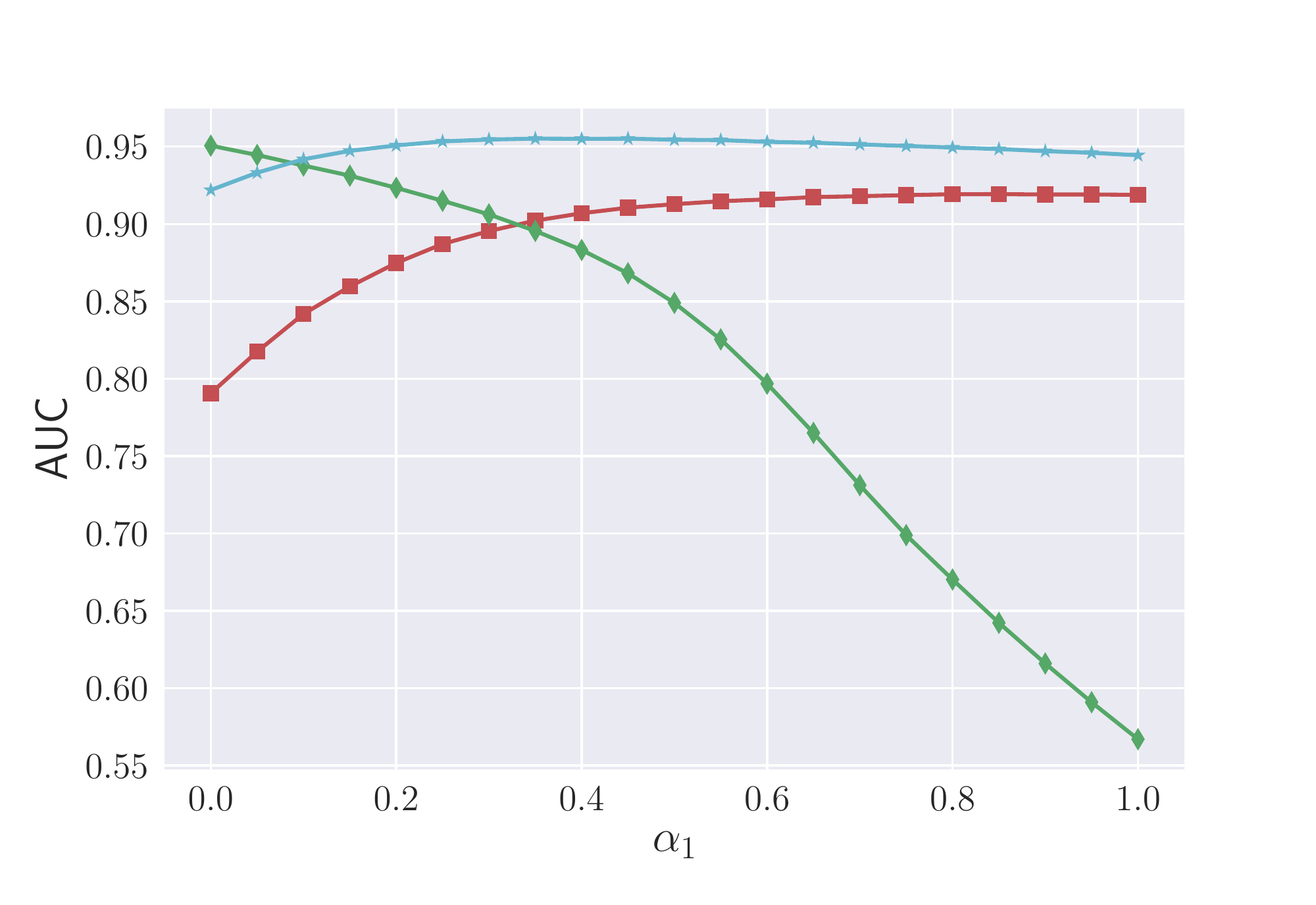} 
 \caption{AUC values as a function of $\alpha_1$ when the HIF score is evaluated according to Eq. \ref{eq:reduced_aggregation} (HIF1), for the 'red' (square), the 'green' (diamond) and  'cyan' (star) anomalies.}
 \label{fig:auc_alpha1}
\end{figure}

This first finding is confirmed by the ROC curves presented in Fig.\ref{fig:hif_roc} obtained for the best $\alpha_1$ value for HIF1. 
%These curves have been evaluated with $\alpha2=1$, namely when only a mixture of $s(x,n)$ and $\tilde{s_c}(x)$ scores are used by the HIF. 
If the ROC curves for the peripheral 'red' and 'cyan' anomaly clusters are close for the two algorithms and apparently rather good, for the 'green' anomalies, the ROC curve for IF is very poor (IF will not perform better than a random classifier) while it is quite good for HIF1.            
Table \ref{tab:table_alpha1_ALL}, in which the global AUC values for the two algorithms are reported, supports this claim. In addition, Fig. \ref{fig:auc_alpha1} visualizes the variation of these AUC values when varying parameter $\alpha_1$ in $[0;1]$. As expected, when $\alpha_1$ increases, the AUC value for the 'green' anomalies regularly decreases, while the AUC values for the 'red' and 'cyan' anomalies increases to reach a plateau when $\alpha_1 > .4$ and $\alpha_1 > .7$ respectively. This explains the optimum value for $\alpha_1$ that corresponds to a compromise between the two scores of different nature that are involved.\\ 

\subsubsection{Impact of the $s_a(x)$ score} We consider at this stage that $\alpha_1=1$ in the HIF score, which leads to the reduced aggregation score given in Eq. \ref{eq:reduced_aggregation_alpha2}.
\begin{equation}
s_{hif2}(x,n) = \alpha_2 \cdot \tilde{s}(x,n) + (1 - \alpha_2) \cdot \tilde{s_a}(x)
\label{eq:reduced_aggregation_alpha2}
\end{equation}

Thus, 5 ($.05\%$) 'red' labeled anomalies are now added into the HIF score.

\begin{figure}[h!]
    \centering
    \includegraphics[width=0.5\textwidth]{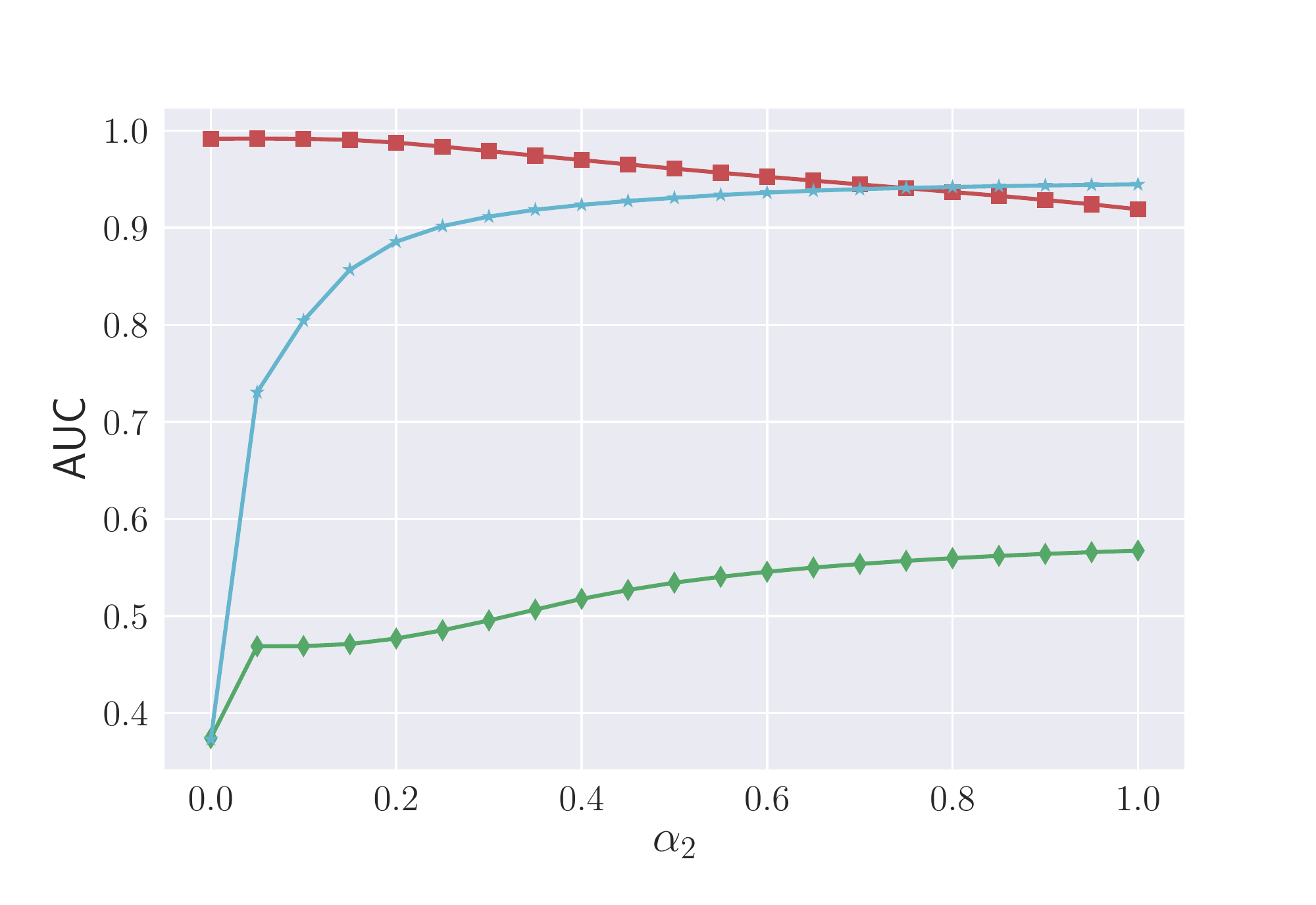} 
 \caption{AUC values as a function of $\alpha_2$ when HIF score is evaluated according to Eq. \ref{eq:reduced_aggregation_alpha2}, for the 'red' (square), the 'green' (diamond) and  'cyan' (star) anomalies.}
 \label{fig:auc_alpha2}
\end{figure}

It can be shown on Fig.\ref{fig:auc_alpha2} that when $\alpha_2$ increases, the AUC value corresponding to the detection of the 'red' anomalies slightly decreases  while it increases from near $.4$ to $.9$ for the 'cyan' anomalies: this is expected, because no 'cyan' labeled anomalies have been added to the HIF. For the 'green' anomalies, when $\alpha_2$ increases, the AUC value increases slightly with mean value near $.5$, basically the chance level. Note that for $\alpha_2=1$ the AUC values correspond to the IF scoring.

\begin{figure}[h!]
    \centering
    \includegraphics[width=0.5\textwidth]{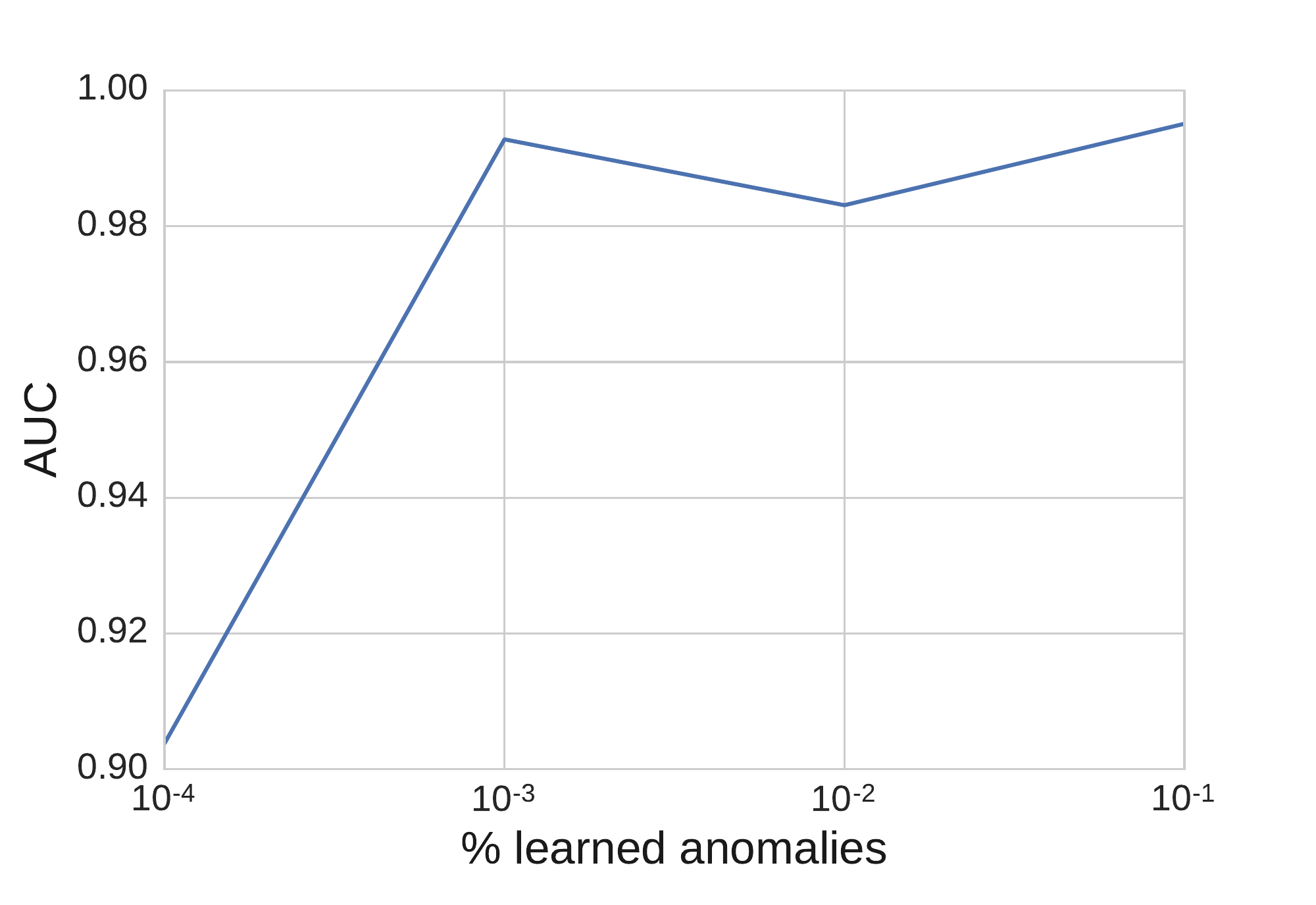} 
 \caption{AUC value as a function of the percentage of 'red' anomalies added into the HIF.  The task consists to separate 'normal' test data from 'red' anomalies.}
 \label{fig:contaminSyn}
\end{figure}

Considering the separation of the 'red' anomalies from the 'normal' test data, Fig. \ref{fig:contaminSyn} gives the best AUC value (obtained when optimizing the meta parameters $\alpha_1$) as a function of the percentage of anomalies (compared to the normal train data) used to train the HIF. We observe that the addition of a single anomaly ($.001\%$)  is enough to increase the AUC from $.9$ to near $.99$. Indeed, to optimize in practice the meta parameters, we will require more labeled anomalies than a single one.

\subsubsection{Evaluating the complete HIF score} We consider here the full HIF score which corresponds to the aggregated scoring given in Equation \ref{eq:aggregation}. We refer to this configuration as the HIF2 algorithm.

\begin{figure}[h!]
    \centering
    \includegraphics[width=0.6\textwidth]{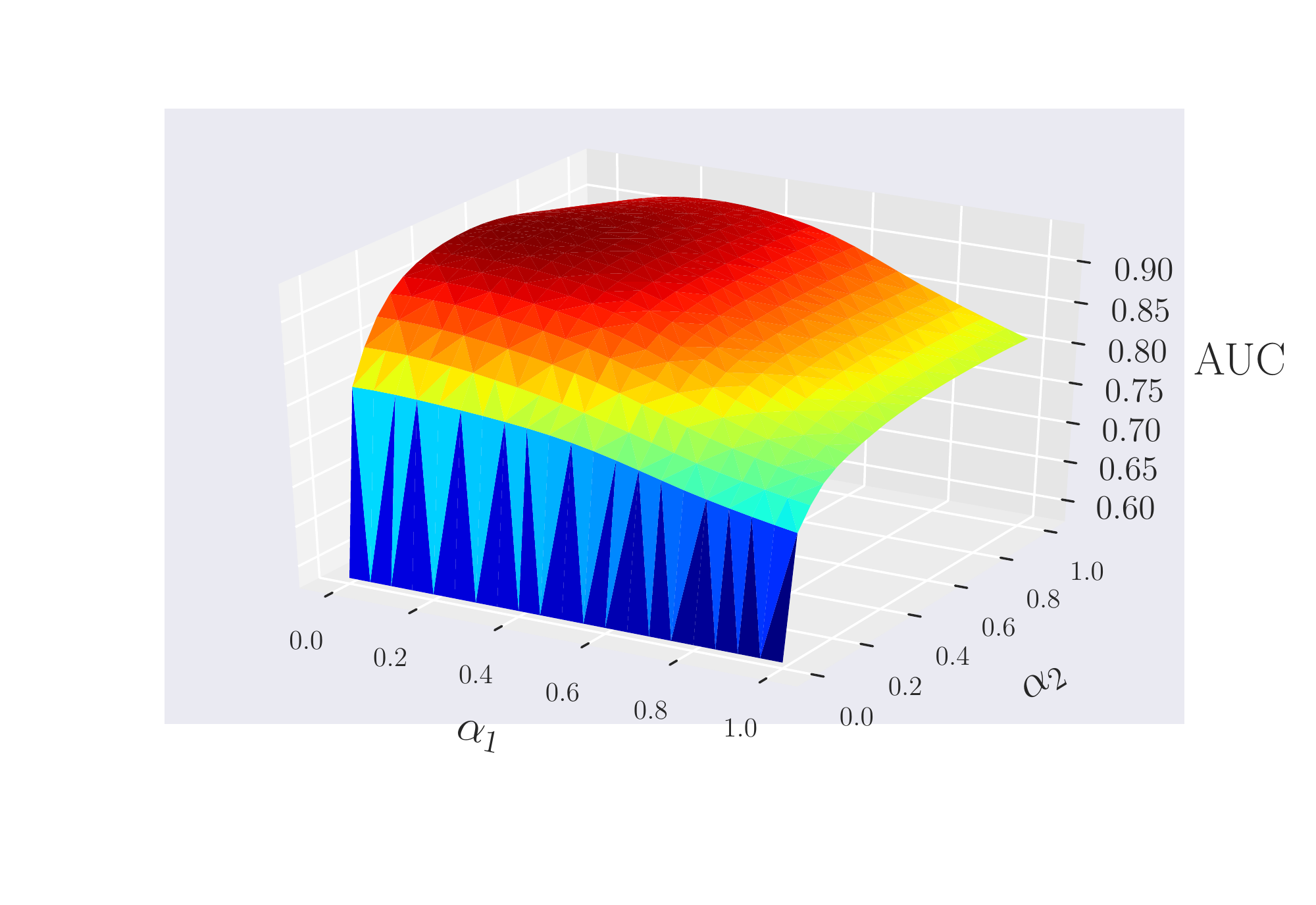} 
 \caption{AUC values as a function of $\alpha_1$ and $\alpha_2$ when HIF score is evaluated according to Eq. \ref{eq:aggregation} (HIF2), and when all the types of anomalies ('red', 'green' and 'cyan') are considered in a single detection test.}
 \label{fig:auc_alpha1_alpha2}
\end{figure}

The 3D plot presented in Fig. \ref{fig:auc_alpha1_alpha2} shows that a maximal AUC value can be obtained when optimizing the two meta parameters $\alpha_1$ and $\alpha_2$.

\begin{table}[h!]
\centering
  \begin{tabular}{l|c||c|c|c|}
    &  IF &  HIF2 (Eq.\ref{eq:aggregation}) & $\alpha_1$ & $\alpha_2$ \\
    \hline\hline
    Mean  & .793  &	.944 & .192 & .71	 \\ \hline
    StdDev & .007 & .008 & .021	& .016	 \\
    \hline
  \end{tabular}
  \caption{Mean best values with standard deviations for $\alpha_1$ and $\alpha_2$, and corresponding AUC values with standard deviations for IF and HIF2 algorithms. For this experiment, 'normal test data are mixed up with all the types of anomalies. }
 \label{tab:table_alpha1_alpha2_ALL}
\end{table}

Table \ref{tab:table_alpha1_alpha2_ALL} presents the mean and standard deviatoion of the AUC values for IF and HIF2 evaluated on ten independent tests (for which the 'normal' and anomaly data are randomly drawn).

A mean optimal AUC value is obtained for $\alpha_1\approx .2$ and $\alpha_2\approx .7$ which means that all the individual scores $s(x,n)$, $s_c(x)$ and $s_a(x)$ play a role. This shows that when some anomalies can be used for training, a simple grid search optimization process can easily be implemented to set up the HIF2 meta parameters $\alpha_1$ and $\alpha_2$.  

The final global AUC values obtained when all the test data ('normal' test, 'red', 'green' and 'cyan' anomalies) are evaluated in a single test are $.793$ for the IF algorithm, $.910$ for the HIF1 algorithm when no labeled anomalies are added, and $.928$ when $0.5\%$ of the red anomalies are added into the HIF2.\\

\begin{figure}[!h]
    \centering
    \begin{tabular}{ccc}
        \centering
        \includegraphics[width=0.32\textwidth]{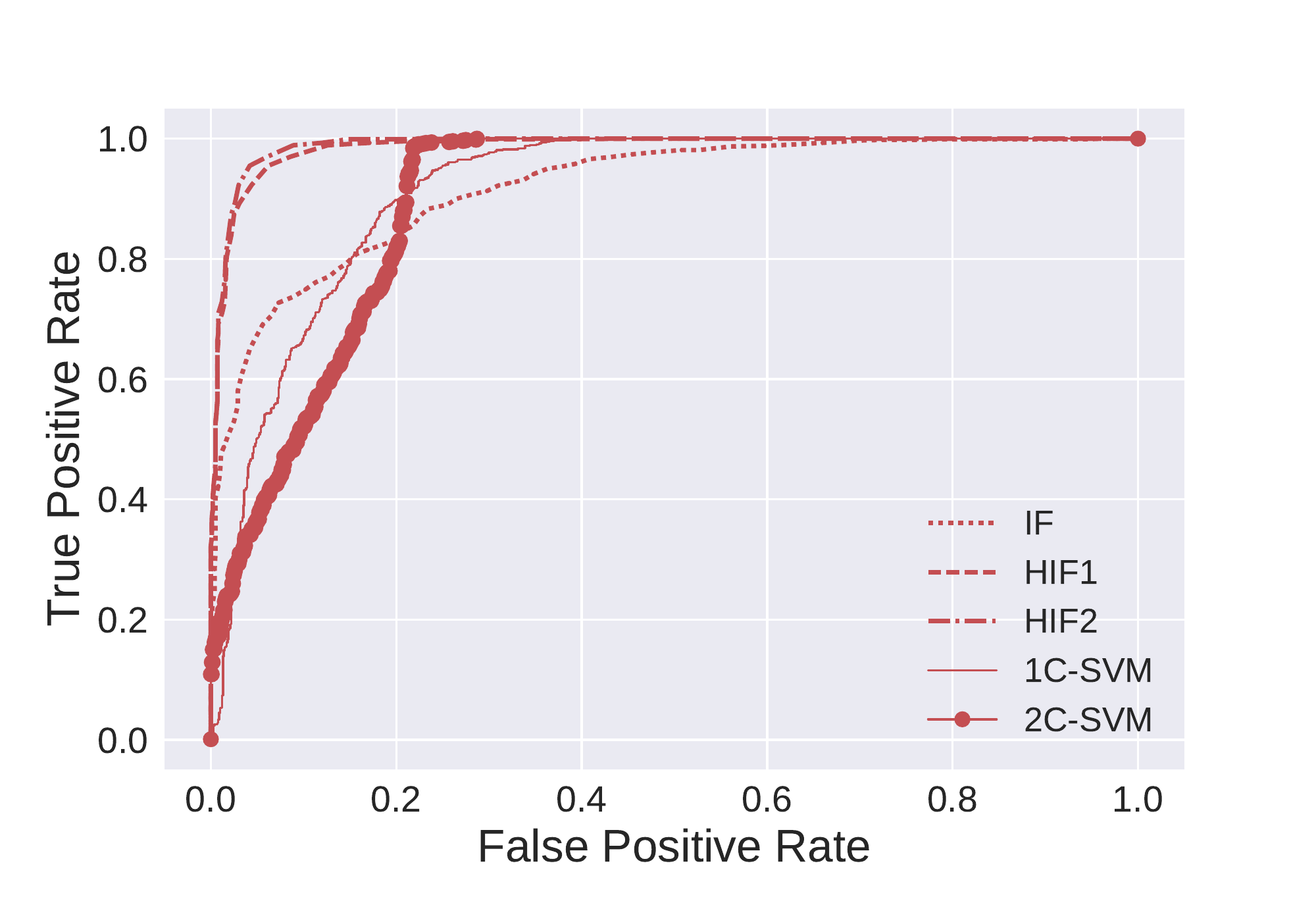} &
        \includegraphics[width=0.32\textwidth]{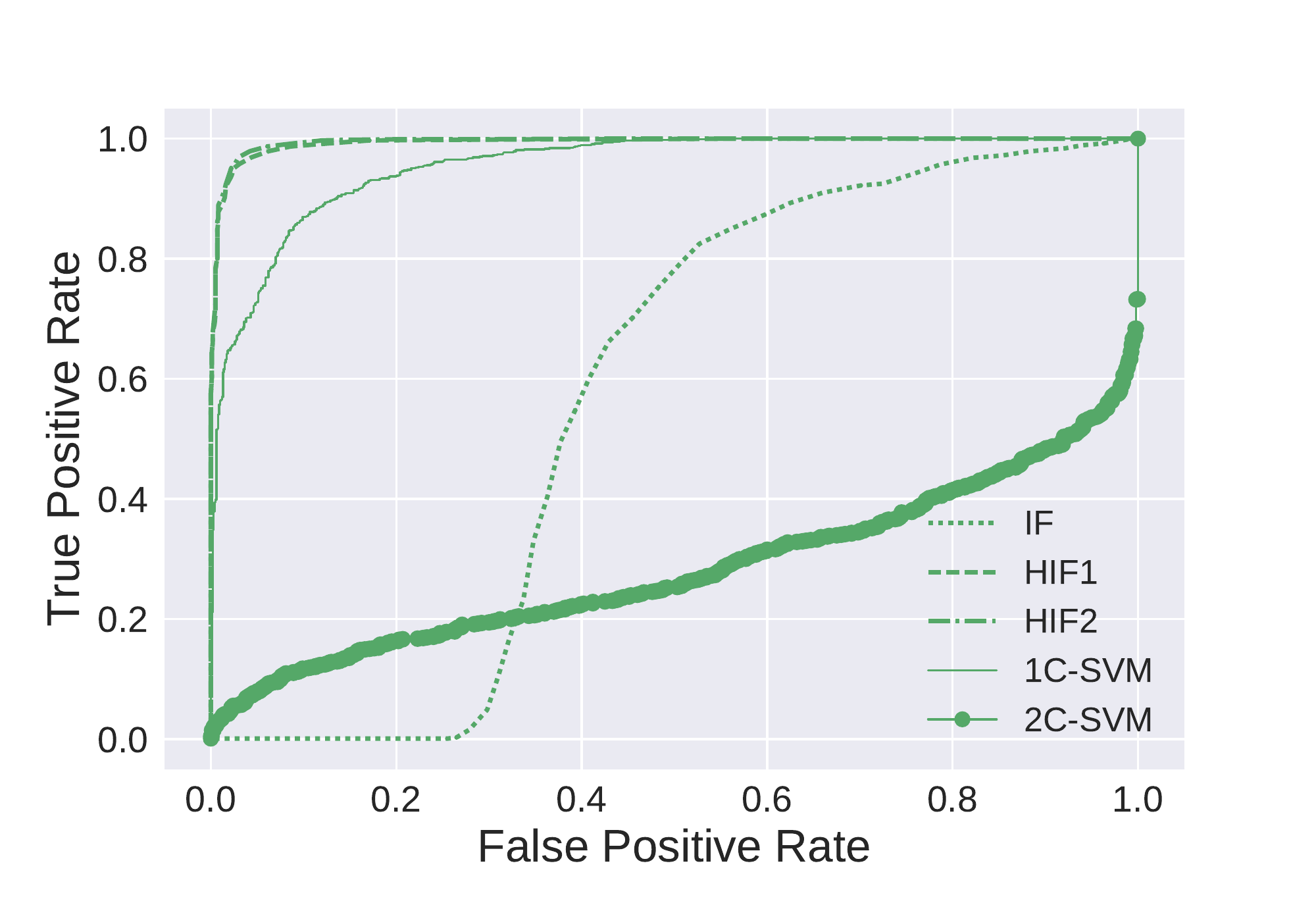} &
        \includegraphics[width=0.32\textwidth]{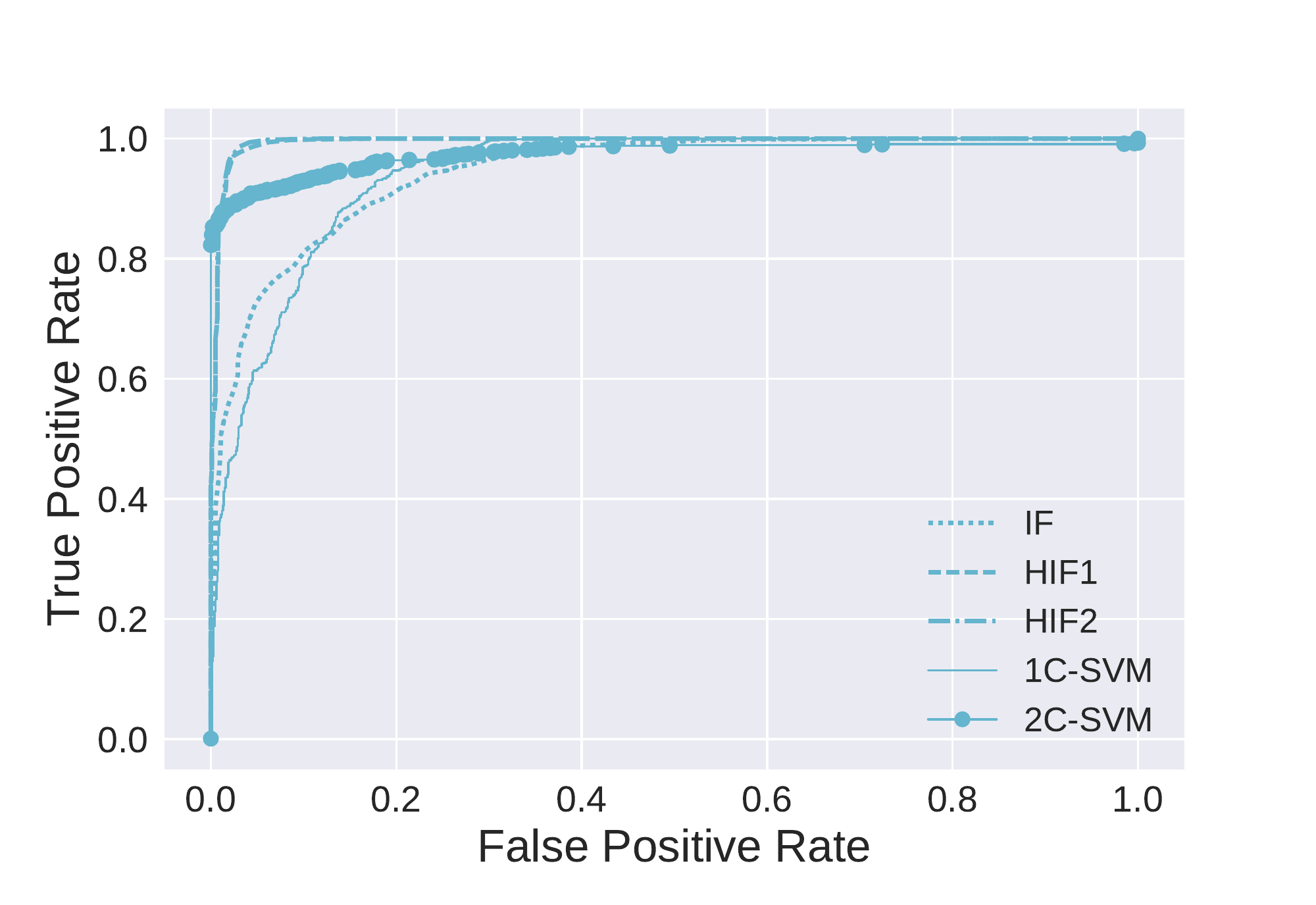}
        \end{tabular}
        \caption{ROC curves for anomaly detectors based on 1C-SCVM (solid line), 2C-SVM (solid circle), IF (dotted), HIF1 (dashed) and HIF2 (dashdot) anomaly scores when the red (left), green (middle) and cyan (right) anomalies are considered in separated tests. }
        \label{fig:hif_roc}
    \end{figure}        
    
As a preliminary conclusion, according to this synthetic experiment, we can state that the 'blind spot' drawback of the IF algorithm has been apparently correctly solved by introducing a distance-to-normality based paradigm into the scoring of the algorithm. Furthermore, adding some few labeled anomalies into the HIF structure provides a supervised complementary source of potential improvements.

\subsubsection{Comparison with the one-class and two-classes SVM}
In order to assess the IF and HIF (HIF1 and HIF2) algorithms comparatively to the state of the art in anomaly detection, we consider here the one-class Support Vector Machine (1C-SVM), and the supervised two-classes Support Vector Machine (2C-SVM) as two baselines. We constructed 15 random instances of the synthetic dataset.  We have selected for the 1C-SVM and the 2C-SVM the radial basis kernel that is very well adapted for separating such kind of non linearly separable uniformly and normally distributed data. On each test, we optimized the ($\nu$ and $\gamma$) meta parameters of the SVM such as maximizing the AUC value of the ROC curve obtained when using the distance to the separating hyper-plane as the classification score for the SVM. The same number of randomly drawn $5$ 'red' anomalies have been used to train the HIF2 and the 2-classes SVM. IF and HIF structures are built using $\psi=64$ and $t=2048$. Similarly to the SVM, we optimize for HIF1 and HIF2 $\alpha_1$ and $\alpha_2$ meta parameters such as maximizing the AUC value that is reported in the following tables.

\begin{table}[h!]
\centering
\caption{Mean AUC values and their standard deviation obtained by the 1C-SVM, 2C-SVM, IF, HIF1 and HIF2 algorithms}\begin{tabular}{|l|l|l|l|}
\hline
 &  Mean AUC & Std. dev. \\ \hline
\hline 
1C-SVM & 0.940 & 0.008   \\ \hline
2C-SVM & 0,870 & 0,011   \\ \hline
 IF & 0.730 & 0.007    \\ \hline
 HIF1 & 0.937  & 0.009   \\ \hline
 HIF2 & 0.944  & 0,008   \\ \hline
\end{tabular}
\label{tab:1CSVMCompare}
\end{table}

\begin{table}[h!]
\centering
\caption{ Wilcoxon signed-rank test of pairwise accuracy differences for 1C-SVM, 2C-SVM, IF, HIF1 and HIF2 classifiers carried out on the 20 synthetic tests. }
\begin{tabular}{|l|l||l|l|l|}
\hline
Method	           &  2C-SVM & IF & HIF1 & HIF2 \\ \hline
\hline 
1C-SVM & \textbf{\texttt{0.0007}} & \textbf{\texttt{0.0007}} & \texttt{0.306} & \texttt{0.139}  \\ \hline
2C-SVM & -&\textbf{\texttt{0.0007}} &\textbf{\texttt{0.0007}} & \textbf{\texttt{0.0007}}   \\ \hline
 IF & -& - & \textbf{\texttt{0.0007}}  & \textbf{\texttt{0.0007}}  \\ \hline
 HIF1 & - & -  & - & \textbf{\texttt{0.0008}}  \\ \hline
\end{tabular}
\label{tab:significance_test}
\end{table}

Table \ref{tab:1CSVMCompare}  presents the mean AUC values and associated standard deviation obtained for each algorithm. Table \ref{tab:significance_test} gives the p-values of the Wilcoxon signed rank tests carried out to evaluate the significance of the pairwise AUC differences for each pair of classifiers. This experiment shows that the 1C-SVM and the HIF1 algorithms perform very comparatively since their pairwise differences are not significant. On the other hand the supervised functionality of HIF2, when adding randomly .5\% of labeled anomalies (namely 5 among 1000 samples), brings a slight improvement compared to the HIF1 algorithm and the 1C-SVM (although the difference between HIF2 and 1C-SVM seems to be not significant here). Nonetheless, the 2C-SVM performs significantly worse than the 1C-SVM and HIF1 and HIF2. According to Fig. \ref{fig:hif_roc}, the 2C-SVM, while learning the separation frontier between the normal and red anomaly data, looses the ability to identify outliers in the center of the tore.

\subsubsection{Dependence to the sample size ($\psi$) in each tree and to the number of trees ($t$) in the forest}
\begin{figure}[h!]
    \centering
     \subfigure[]{
        \includegraphics[width=0.5\textwidth]{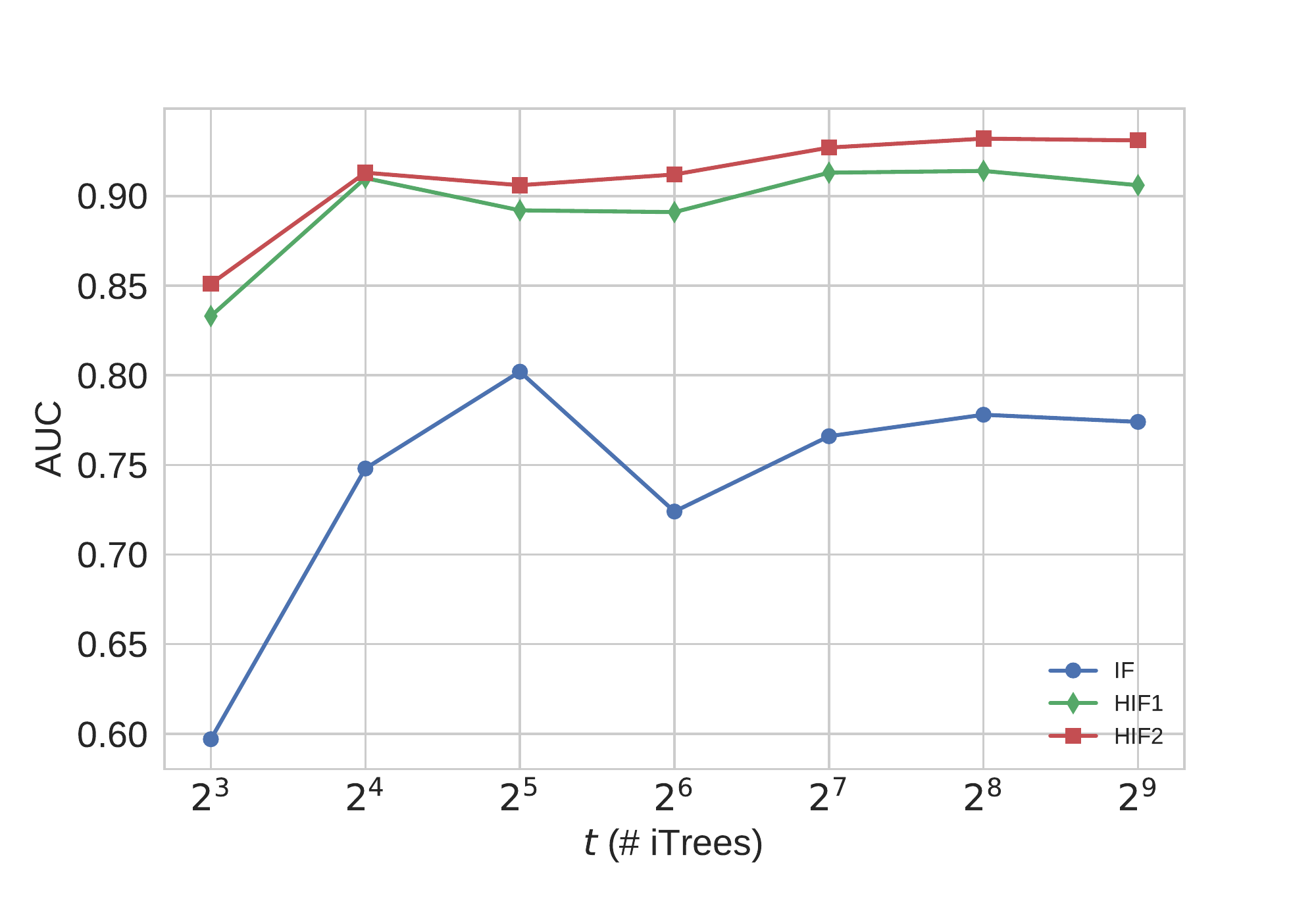}  
            }
    ~
     \subfigure[]{
        \includegraphics[width=0.5\textwidth]{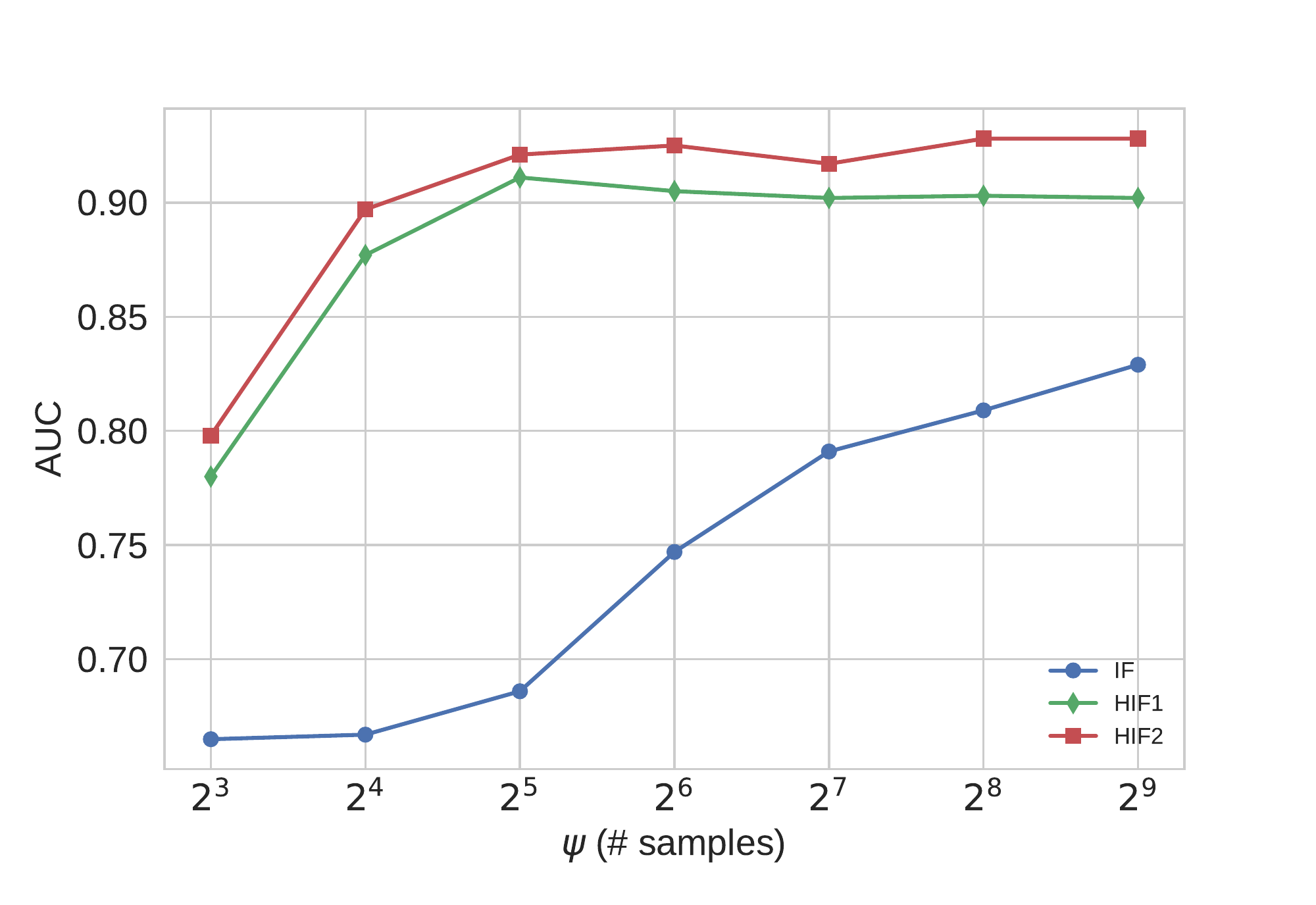}  
    }
 \caption{AUC values: (a) when the number of trees varies while the sample size remains constant equal to 128 samples; (b) when the sample size varies while the number of trees in the forest remains constant equal to 128 trees.}
\label{fig:meta_parameters}
\end{figure}

: The dependence of the AUC values to the other meta parameter settings, namely the number of iTrees and the sample size assigned to each iTree, is presented in Fig \ref{fig:meta_parameters}.

For this experiment that is characterized with a dataset size of $n=1000$ samples used to train the isolation forest, one can see that the HIF reaches good and relatively stable AUC values with fewer trees ((a) sub-figure) and lower sample size ((b) sub-figure) which is quite advantageous in terms of memory space and response time.

\subsection{Setting up meta parameters $\alpha_1$ and $\alpha_2$}
\label{sec:meta_alpha1_alpha2}
We consider here that 'normal' data are available for training.

\subsubsection{No labeled anomaly is available}: in this case we only need to setup $\alpha_1$. Figures \ref{fig:auc_all_alpha1} and \ref{fig:auc_alpha1} shows that a default value for $\alpha_1$ can be setup in $[.2;.5]$. If numerous unlabeled anomalies exist in the test data as well as 'normal' test data, then an exploratory analysis of the distribution of the HIF1 score may be used to isolate anomaly modes from the 'normal' distribution. By varying $\alpha_1$, the separation of the distribution modes may vary and the seek for an 'optimal' separation may be carried out.

 \begin{figure}[h!]
    \centering
    \includegraphics[width=0.6\textwidth]{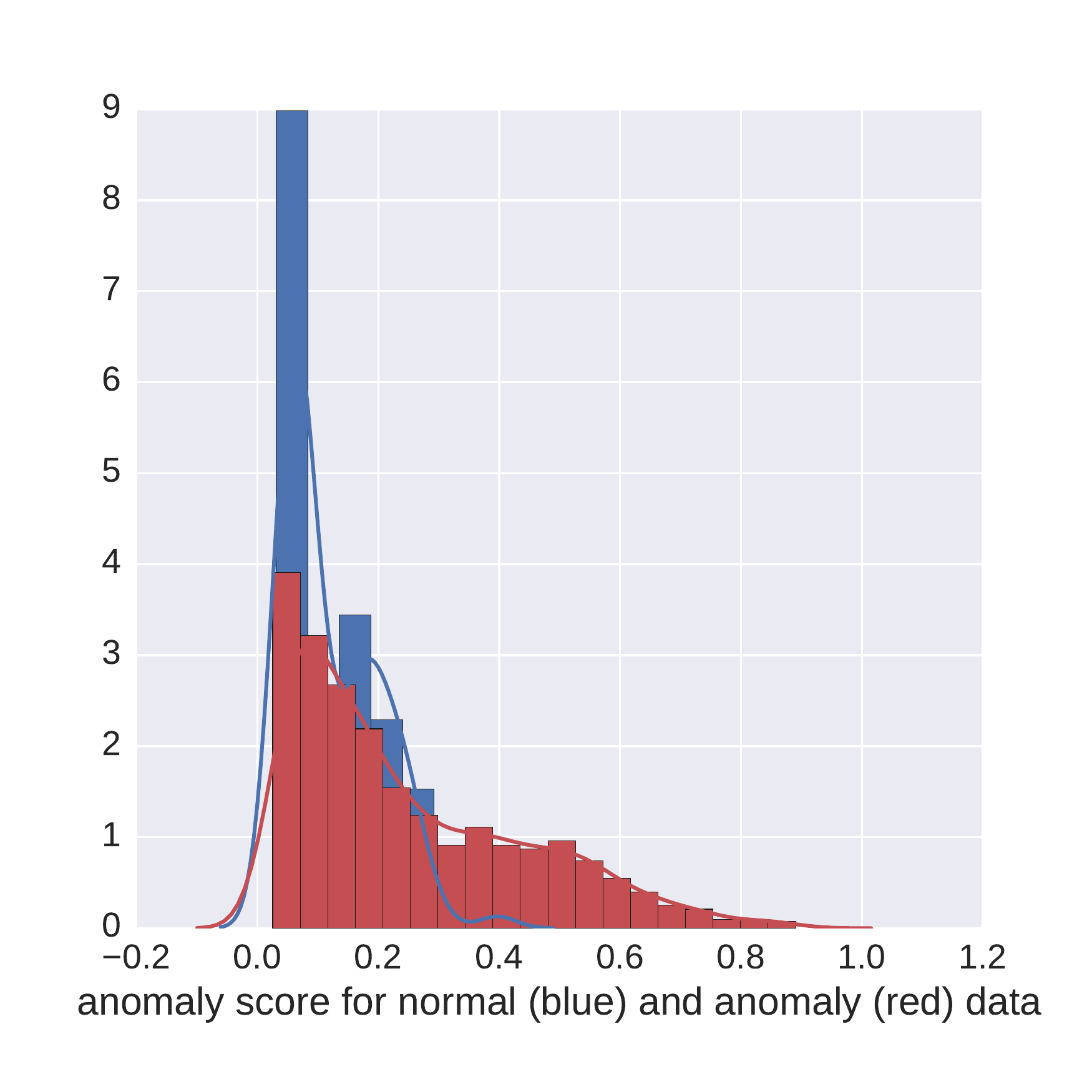} 
 \caption{Anomaly scores when 'normal' and 'anomaly' data are mixed up in the train data used to build the IF or HIF.}
 \label{fig:distrib_EA}
\end{figure}

As an example, Fig. \ref{fig:distrib_EA} shows, for $\alpha_1=.4$  the distribution HIF1 scores for  'normal' train data distribution (in blue) and the test data (in red) in which 'normal' and 'red' anomalies are mixed up. Clearly, the tail of the test data distribution is much longer towards the high scores than for the train data. This is obviously cues for anomaly identification. Hence, one can sample few data with high score  and expertise them to decide whether they correspond to some kind of anomalies, and doing so, we may be able to provide some labeled anomaly data. This is the basis for designing an some active learning process. 

\subsubsection{Few labeled anomalies are available}: in that case, labeled anomalies can be used in a supervised learning process. A simple grid search can be setup to optimize the detection of the anomalies as we did in our previous test to estimate the impact of the HIF additional scores.\\

If several distinct kind of anomalies exist, an iterative procedure can be set up to progressively isolate the anomalies according to their distinct patterns.
\subsection{Complexity of the HIF algorithm}
Basically, the HIF algorithm has the same overall complexity than the IF algorithm, although some extra computational costs are involved during training and testing stages.

IF has time complexities of $O(t\cdot\psi \cdot log(\psi))$ in the training stage and $O(n \cdot t \cdot log(\psi))$ in the testing stage.

\begin{figure}[h!]
    \centering
    \includegraphics[width=0.6\textwidth]{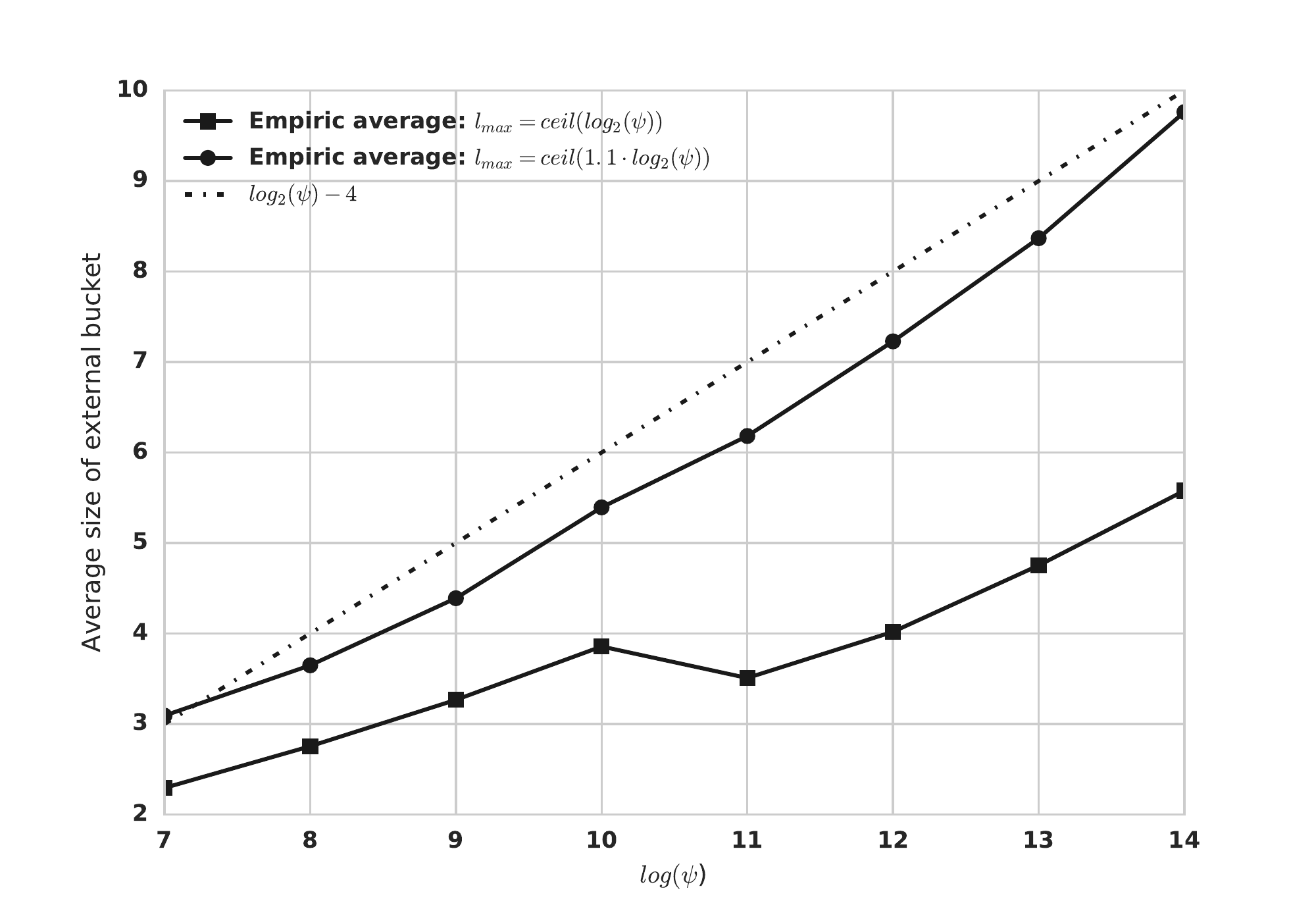}  
 \caption{Average number of instances contained in the external nodes of the iTrees as a function of $log_2(\psi)$; when the maximal height of the iTrees is $l_{max}=\ceil{log_2(\psi)}$ (dotted line),  $l_{max}=\ceil{1.1\cdot log_2(\psi)}$ (circle markers) and $l_{max}=\ceil{1.2\cdot log_2(\psi)}$ (square markers).}
\label{fig:bucketSize}
\end{figure}

At training stage, the HIF algorithm requires to evaluate the centroids of the data attached to each of the external nodes, hence $\psi$ centroids in average need to be evaluated. The evaluation of a centroid is dependent on the number of elements contained in the external buckets ($n_{eb}$). It seems difficult to estimate precisely the expectation of $n_{eb}$, nevertheless, Fig.\ref{fig:bucketSize} presents the result of an empirical study that shows that if the maximal height of the iTree is $l_{max}=\ceil{log_2(\psi)}$, then the average $n_{eb}$ value increases slightly faster than a logarithmic law. For this test, the random forest has been build from a normally distributed dataset with $(0.0,0.0)$ mean, and $((3, 0), (0, 3))$ covariance matrix. Hence, to maintain the overall algorithmic complexity at training stage close to $O(ln(\psi))$ one may increase slightly the maximal height of the iTrees. One can use for instance  $l_{max}=\ceil{1.2\cdot log_2(\psi)}$ that empirically maintains a sub-logarithmic growth a shown in Fig.\ref{fig:bucketSize} . 

Furthermore, at training stage, the HIF2 algorithm requires to evaluate the centroids of the anomalies that are attached to each of the external nodes. As in general very few anomalies are introduced into the HIF, this extra cost is marginal compared to the others.

At testing time, HIF has the same 'big-O' complexity than the IF, namely $O(n \cdot t \cdot log \psi)$, although some extra distances to centroid costs are involved in HIF. The increase of $l_{max}$ will increase slightly the number of comparisons during the test stage, but this number will still be proportional to $log(\psi)$.

In the following application we have used $l_{max}=\ceil{1.1\cdot log_2(\psi)}$.

\section{Application to intrusion detection in network systems}
We here present the results of the experiments performed. We first describe the dataset retained to conduct our experiments, then specify the pre-processing procedure of data, prior to present and discuss about the results obtained.

In this research work, we are interested in the packet payloads as well as the packet header information. While packet headers generally constitute only a small part of whole network traffic data, packet payloads are more complicated. Accordingly, packet payloads analysis is more costly  than the analysis of packet header data as it needs more computations and pre-processing. 
In this way, considering a suitable pre-processing process of network traffic data is vital.

\subsection{The ISCX dataset}
The ISCX dataset 2012 \cite{Shiravi:2012:TDS:2622690.2623143}, which has been prepared at the Information Security Centre of Excellence at the University of New Brunswick, is used to perform this experiments and evaluate the performance of our proposed approaches.
The entire ISCX labeled dataset comprises over two million traffic packets characterized with 20 features and covered seven days of network activities (i.e. normal and attack).
%Real traces generated real traffic for HTTP, SMTP, SSH, IMAP, POP3, and FTP application layers.
Four different attack types, called as Brute Force SSH, Infiltrating, HTTP DoS, and DDoS are conducted on different days. 
Despite some minor disadvantages, the ISCX dataset 2012 is the most up-to-date available dataset compared to the other aging explored datasets for intrusion detection \cite{AKYOL20162015EDP7357,DBLP:journals/sj/LinLWCL16,Shiravi:2012:TDS:2622690.2623143}.

As input to the detection process, we make use of the pre-processed ISCX flows consisting of header and payload packets which will be detailed later. To do so, first of all, flows are classified according to their application layers types such as HTTP Web, POP, IMAP, SSH, etc. Then we do the experiments separately on each data subset.  
This choice has been adopted because the 'normal' traffic flows look very different depending on the application or service layer they relate to. It is also a way to reduce the complexity of the supervision task.

\subsection{Pre-processing of the data}
Data pre-processing is a crucial task which can be even considered as a fundamental building block of intrusion detection. 
Pre-processing involves cleaning  the data and removing redundant and unnecessary entries. It also involves converting the features of the dataset into numerical data and saving in a machine-readable format.
To  convert categorical features into entirely numerical ones, we adopt the binary number representation where we use $m$ binary numbers to represent a $m$-category feature.
However, when a categorical feature takes its values in an infinite set of categories, we need to consider another conversion approach. To do so, we use histogram of distributions. 

In the next step, we add the number of source-destination pairs in a pre-defined window size of flows to the features. This added feature helps to detect network and IP scans as well as distributed attacks. Table \ref{tab:features} presents the final features after pre-processing which be used in the experiments.

\linespread{0.6}
\begin{table}[]
\setlength{\tabcolsep}{10pt}
\centering
\caption{Feature names}
\label{tab:features}
\begin{tabular}{@{}cccc@{}}
\hline
\# of feature & feature name   & \# of feature & feature name         \\ \hline
1             & dest Payload0  & 26            & protocol Name4       \\
2             & dest Payload1  & 27            & protocol Name5       \\
3             & dest Payload2  & 28            & source Payload0      \\
4             & dest Payload3  & 29            & source Payload1      \\
5             & dest Payload4  & 30            & source Payload2      \\
6             & dest Payload5  & 31            & source Payload3      \\
7             & dest Payload6  & 32            & source Payload4      \\
8             & dest Payload7  & 33            & source Payload5      \\
9             & dest Payload8  & 34            & source Payload6      \\
10            & dest Payload9  & 35            & source Payload7      \\
11            & dest Port      & 36            & source Payload8      \\
12            & dest TCPFlag0  & 37            & source Payload9      \\
13            & dest TCPFlag1  & 38            & source Port          \\
14            & dest TCPFlag2  & 39            & source TCPFlag0      \\
15            & dest TCPFlag3  & 40            & source TCPFlag1      \\
16            & dest TCPFlag4  & 41            & source TCPFlag2      \\
17            & dest TCPFlag5  & 42            & source TCPFlag3      \\
18            & direction0     & 43            & source TCPFlag4      \\
19            & direction1     & 44            & source TCPFlag5      \\
20            & direction2     & 45            & duration             \\
21            & direction3     & 46            & total dest Bytes     \\
22            & protocol Name0 & 47            & total dest Packets   \\
23            & protocol Name1 & 48            & total source Bytes   \\
24            & protocol Name2 & 49            & total source Packets \\
25            & protocol Name3 & 50            & \# of pairs IP       \\ \hline
\end{tabular}
\end{table}
\linespread{1}

Last step, but certainly not the least one, is to normalize the data.
This step is crucial when dealing with features of different units and scales.
Without normalization, features with extremely greater values dominate the features with smaller values.
Here we use min-max normalization approachaccording to which we fit all the features into the unit segment $[0;1]$.

\subsection{Results}

We compare in Table \ref{tab:aucISCX} the AUC values obtained, for the various tested application layers, IF, HIF1 and HIF2 algorithms and the two baselines 1C-SVM and 2C-SVM. In this table, \#LA column gives the absolute number and Labeled Anomalies that are introduced in the HIF2 and used to train the 2C-SVM (Eq.\ref{eq:aggregation}). In this same column, the value in parenthesis corresponds to the total number of anomalies observed for the given application layer. For IF, HIF1 and HIF2, the forests comprise $1024$ trees and each tree is associated to a data sample containing $512$ instances.  For HIF and SVM approaches, the meta parameters have been selected such as maximizing the AUC values. 

The obtained AUC values are in general relatively high, except for the HTTPImage transfer application layer for which the AUC values for IF, HIF1 and HIF2 are respectively $.55$, $.56$ and $.60$. For this application layer, the detection of anomalies are much difficult because the attacks are deeply encoded into the image data packets that are transferred. The features we are using to represent the network data flows are not sufficiently discriminant to detect such anomalies. We note that the two baselines perform slightly better than the isolation forest based algorithms since 1C-SVM and 2C-SVM get respectively $.64$ and $.69$ AUC values for this application layer. 

\begin{table}[h!]
  \centering
  \caption{AUC values for the tested application layers and the tested methods. \#LA is the absolute number of Labeled Anomalies introduced in HIF2 (Eq. \ref{eq:aggregation}).}
  \label{tab:aucISCX}
  \begin{tabular}{l|c|c||c||c|c||c}
    \hline
    application layer & 1C-SVM  & 2C-SVM &IF & HIF1 (Eq. \ref{eq:reduced_aggregation}) & HIF2 (Eq.\ref{eq:aggregation} ) & \#LA (\#A) \\
	\hline \hline
    HTTPWeb & .8271 & .9514 & .8618 & .8618 & .9920 &  400 (40351)\\
    \hline
    HTTPImageTransfer & .6386 & .6948 & .5529 & .5606 & .6048 &  7 (64)\\
    \hline
    POP & .9475 & .9655 & .9445 & .9617 & .9891 & 9 (96)\\
    \hline
    IMAP & .9974 & .9955 & .9956 & .9956 & .9965 & 9 (134)\\
    \hline
    DNS & .6873 &  .8254 & .8280 & .8313 & .8411 & 7 (73)\\
    \hline
    SSH & .9790 & .9792 & .9791 & .9791 & .9872 & 70 (7407)\\
    \hline
    SMTP & .9734 & .9917 & .9877 & .9881 & .9939 & 7 (76)\\
    \hline
    FTP & .9549 &  .9971 & .9973 & .9974 & .9996 & 9 (226)\\
    \hline
    ICMP & .9311 & .9906 & .9876 & .9885 & .9913 &  9 (295)\\
  \end{tabular}
\end{table}

According to table \ref{tab:aucISCX}, the resolution of the 'blind spots'  seems to improve slightly the AUC values in general, although the significance of these improvements is not always established. Clearly, improvements are mostly driven by the incorporation of some labeled anomalies into the HIF. It is particularly visible for the HTTPWeb application layer for which we have added 400 anomalies representing 1\% of all HTTPWeb anomalies. For this application layer, the AUC values are $.86$ for IF and $.99$ for HIF2 which outperforms largely the two baselines. 

  \begin{figure}[h!]
    \centering
     \subfigure[HTTPWeb]{
        \includegraphics[width=0.3\textwidth]{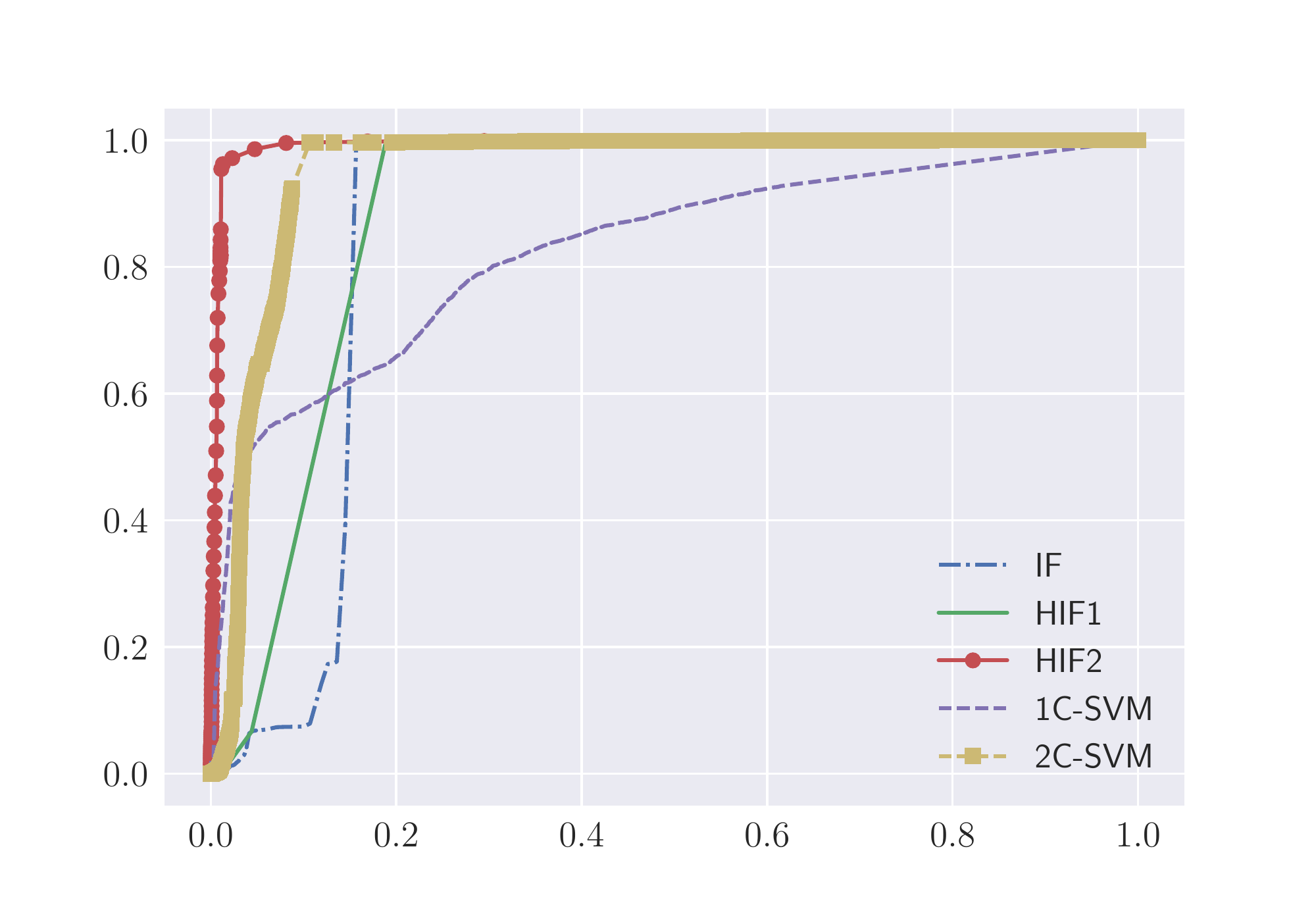}  
        }
    ~
     \subfigure[HTTPImageTransfer]{
        \includegraphics[width=0.3\textwidth]{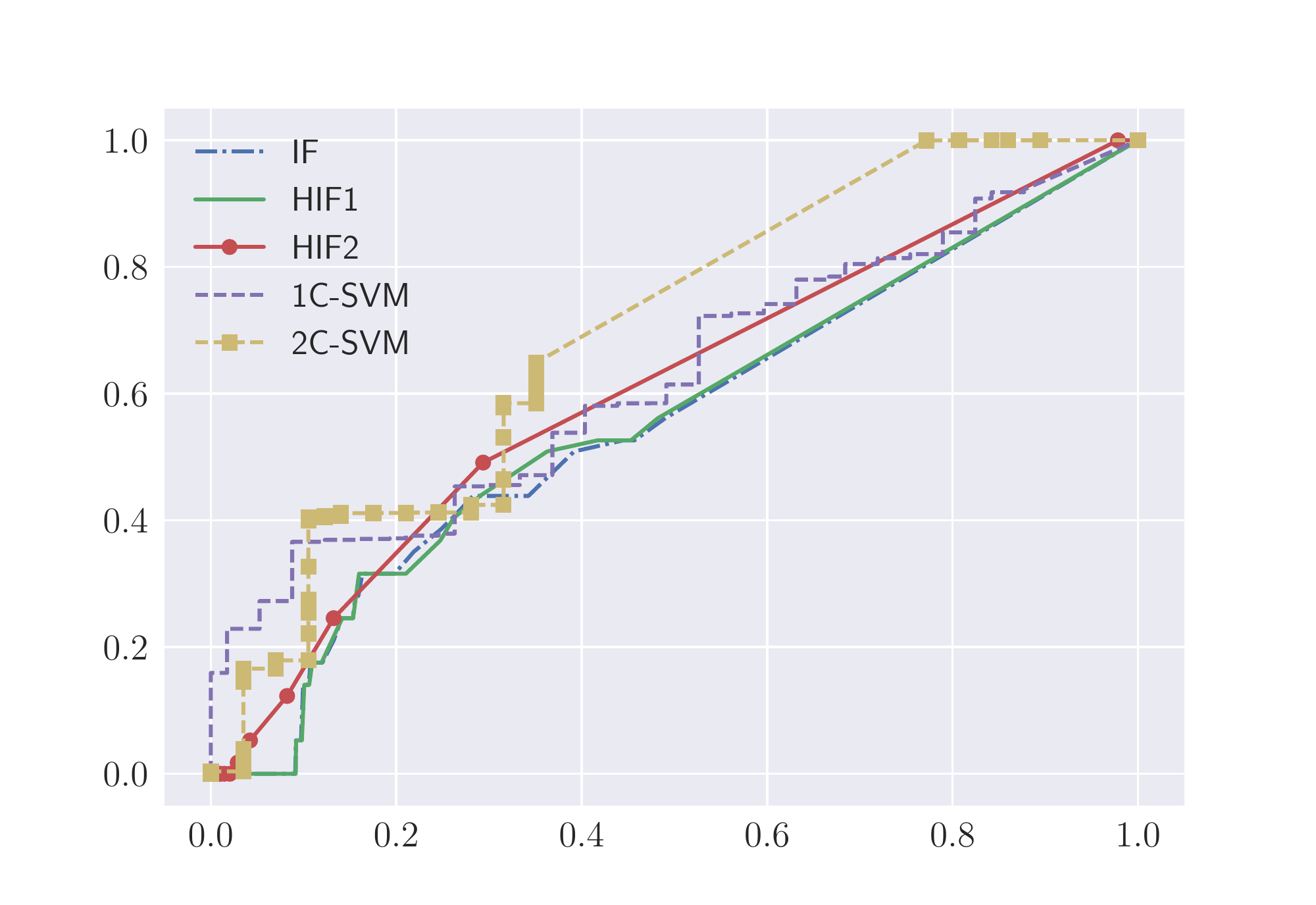} 
        }
    ~
     \subfigure[POP]{
        \includegraphics[width=0.3\textwidth]{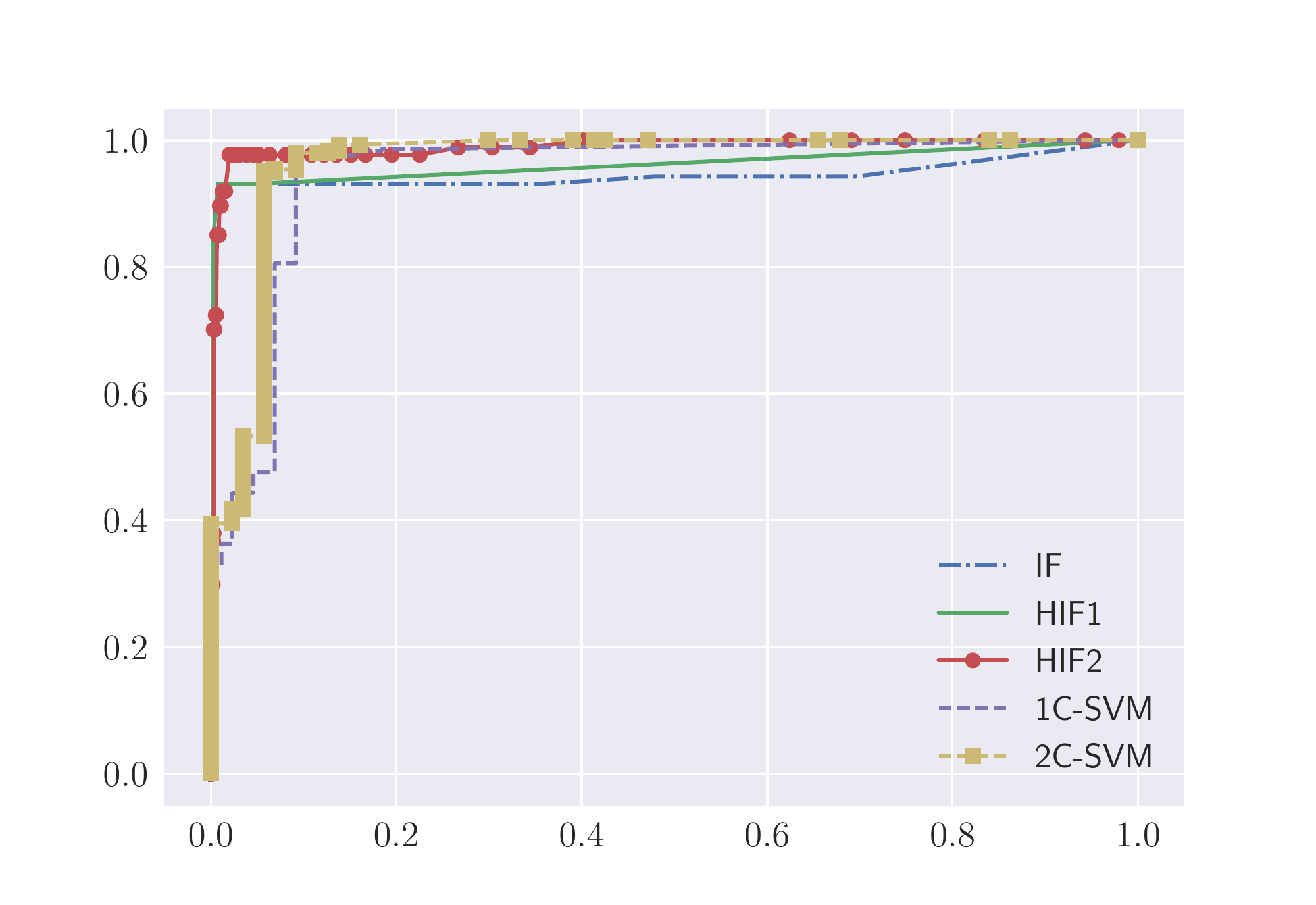}  
        }\\
     \subfigure[IMAP]{
        \includegraphics[width=0.3\textwidth]{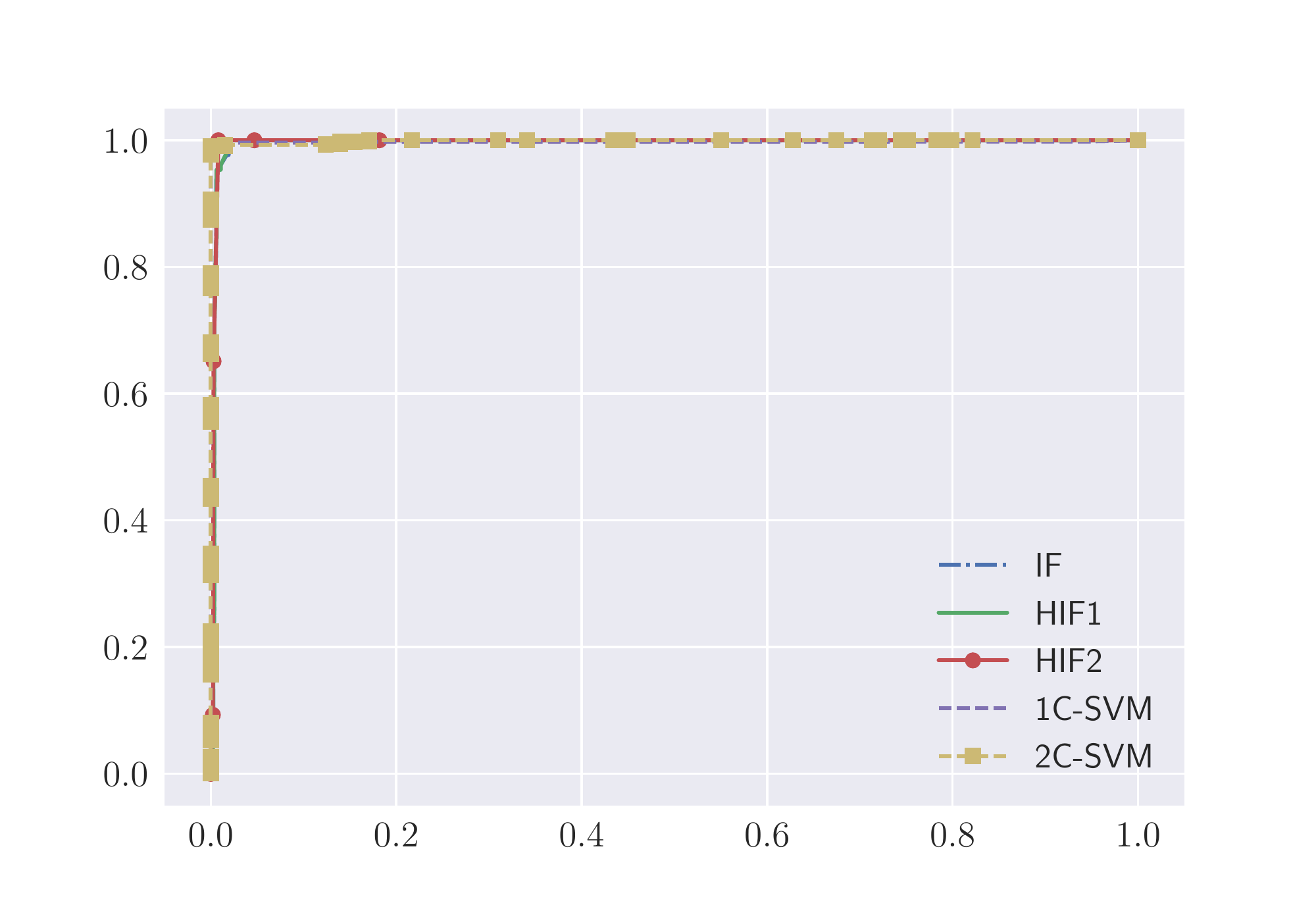}  
        }
    ~
     \subfigure[DNS]{
        \includegraphics[width=0.3\textwidth]{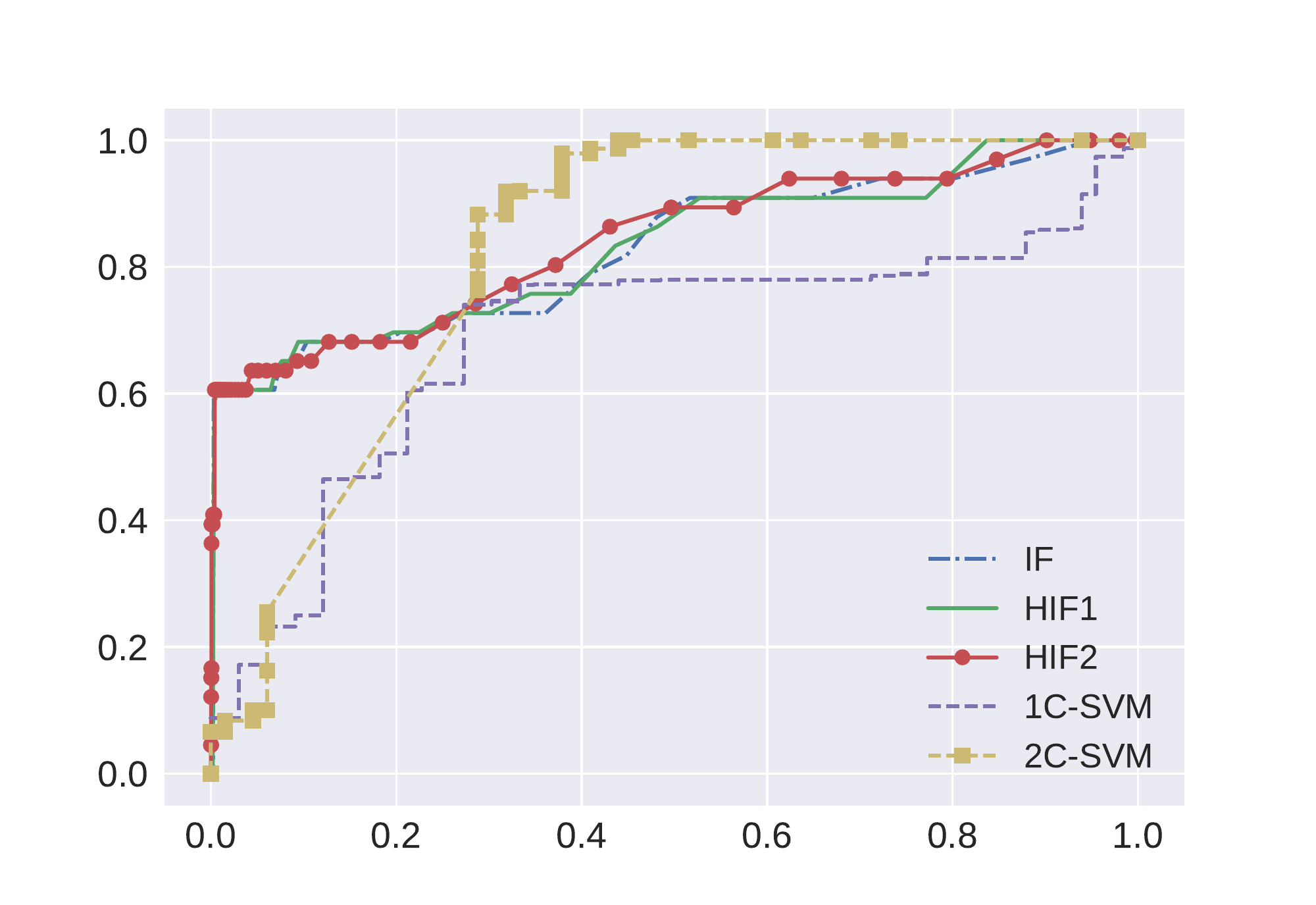} 
        }
    ~
     \subfigure[SSH]{
        \includegraphics[width=0.3\textwidth]{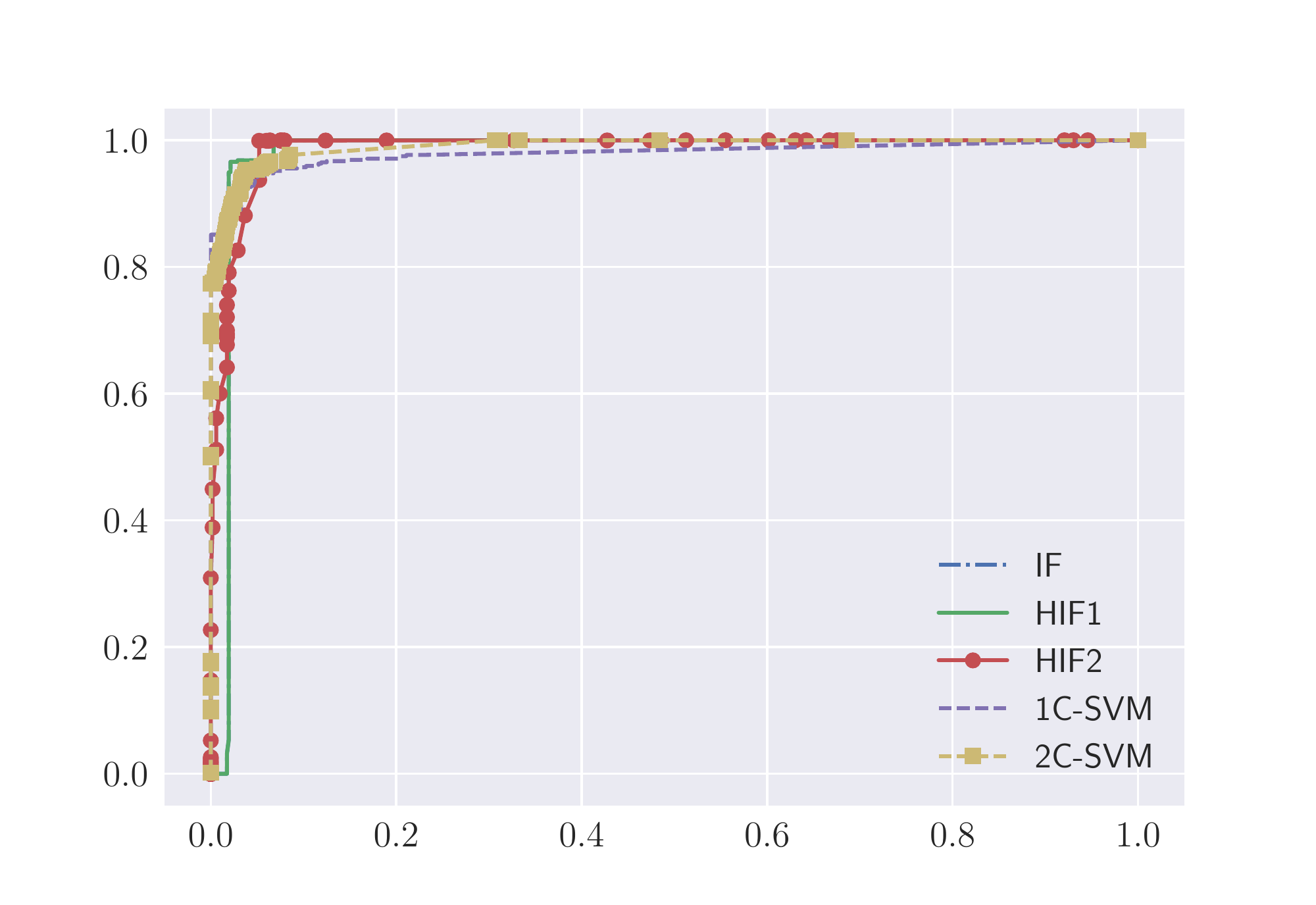}  
        }\\
     \subfigure[SMTP]{
        \includegraphics[width=0.3\textwidth]{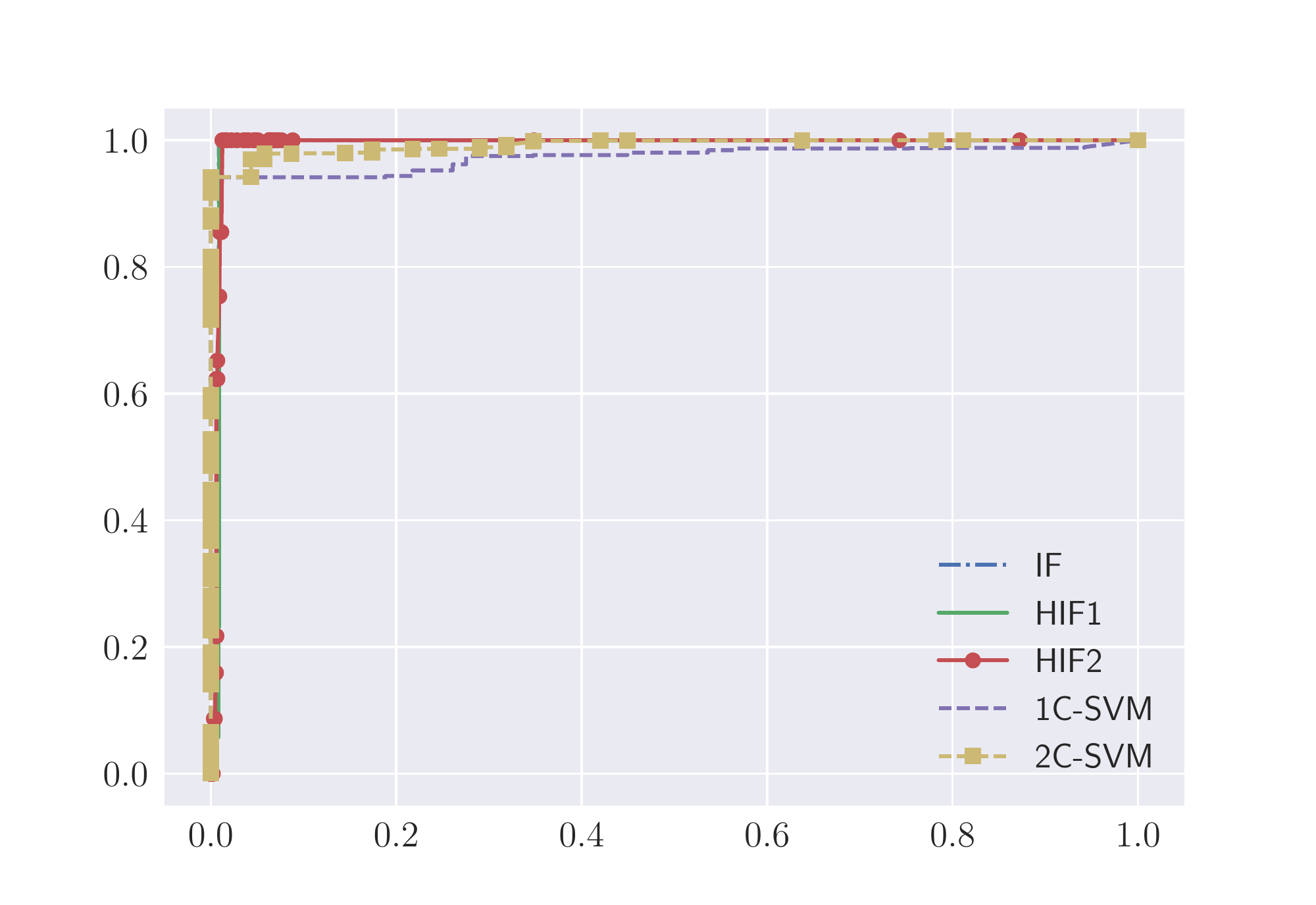}  
        }
    ~
     \subfigure[FTP]{
        \includegraphics[width=0.3\textwidth]{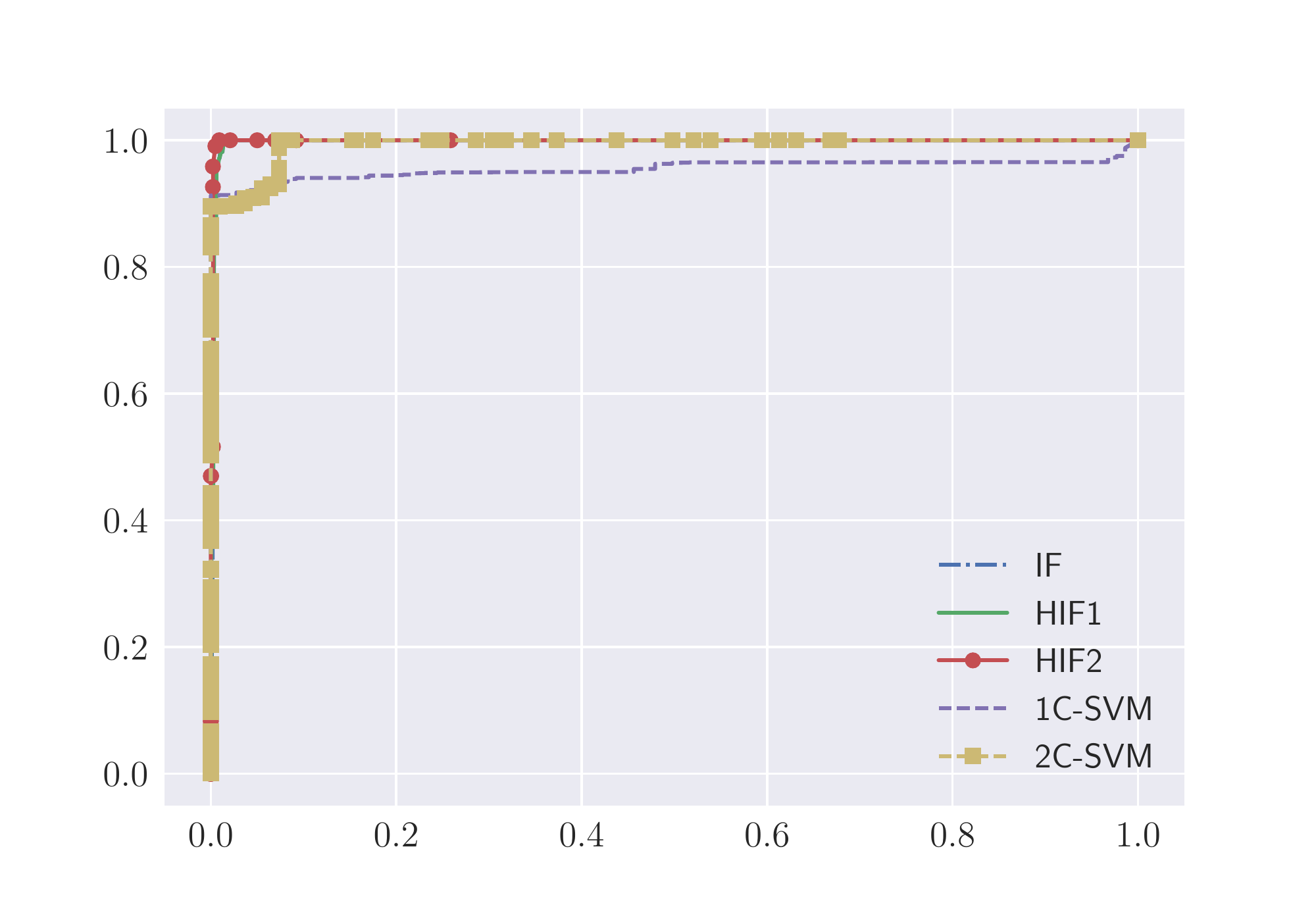} 
        }
    ~
     \subfigure[ICMP]{
        \includegraphics[width=0.3\textwidth]{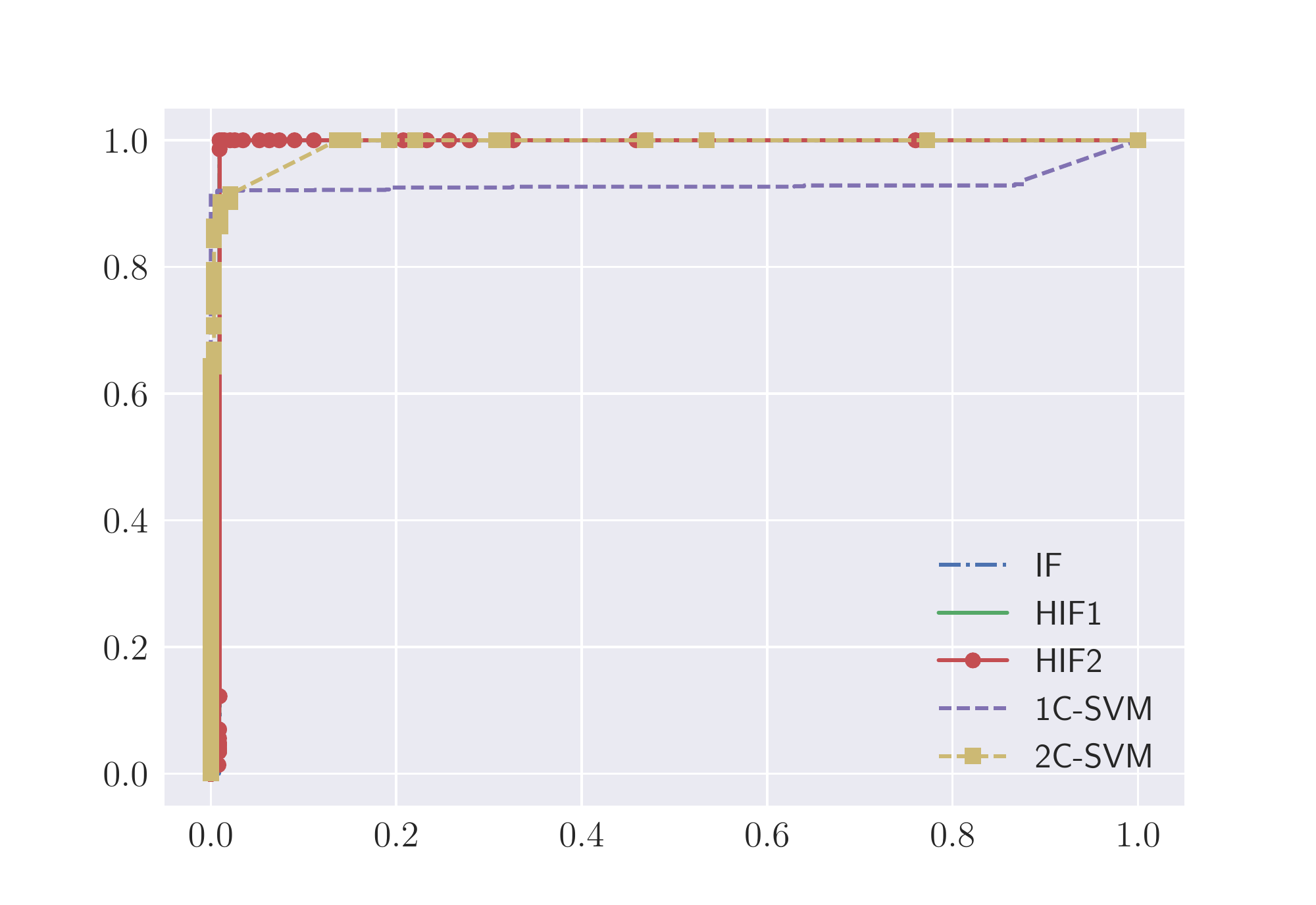}  
        }
 \caption{ROC curves for the nine tested application layers and for the four anomaly detection methods: IF (dash-dotted, blue),  HIF1 (solid, green), HIF2 (circle, red), 1C-SVM (dotted, purple), 2C-SVM (dotted, square, yellow)}
\label{fig:iscx_roc}
\end{figure}

  \begin{figure}[h!]
    \centering
     \subfigure[]{
        \includegraphics[width=0.5\textwidth]{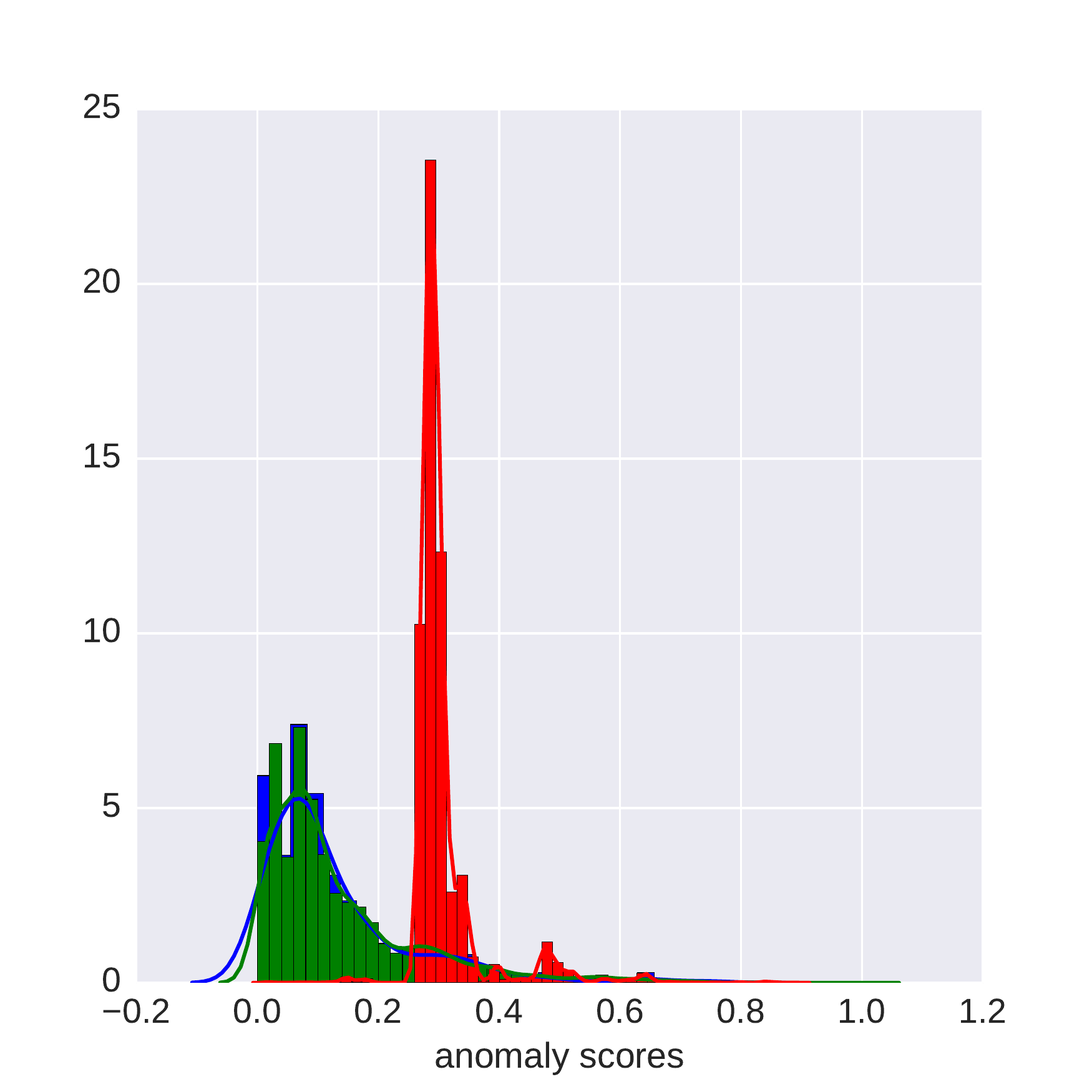}  
            }
    ~
     \subfigure[]{
        \includegraphics[width=0.5\textwidth]{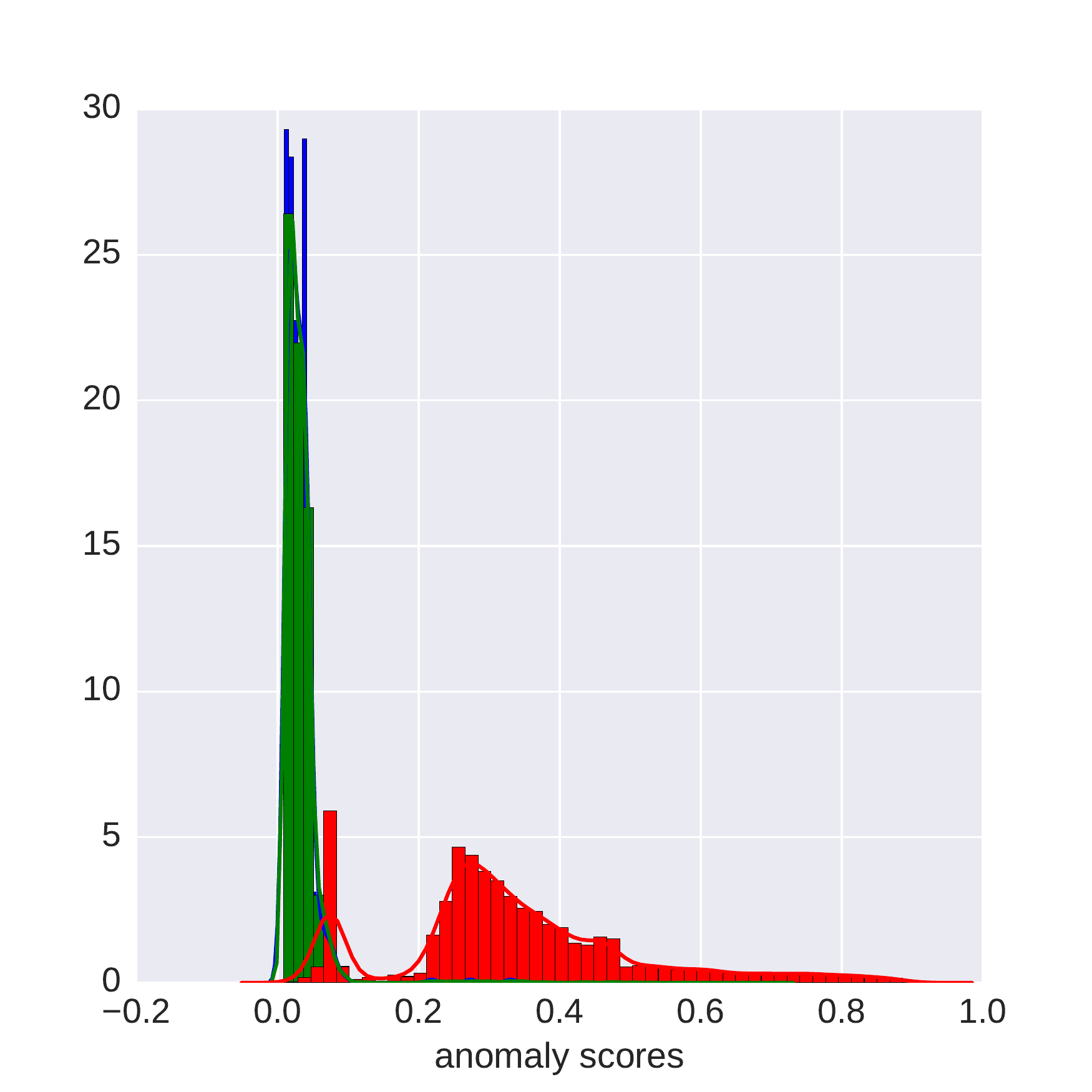}  
    }
 \caption{Detection of anomalies in the ISCX HTTPWeb packet flows: score distributions for train 'normal' data (blue), test 'normal' data (green), and anomalies (red): (a) IF, (b) HIF.}
\label{fig:iscx_httpweb}
\end{figure}

Figure \ref{fig:iscx_roc} shows the ROC curves for the 5 tested methods and the 9 application layers. According to these curves, the HIF1 performs similarly or better than the 1C-SVM and the HIF2 algorithm performs similarly or better than the 2C-SVM, except for the HTTPImageTransfer application layer. 

\begin{figure}[h!]
 \centering
 \includegraphics[width=0.5\textwidth]{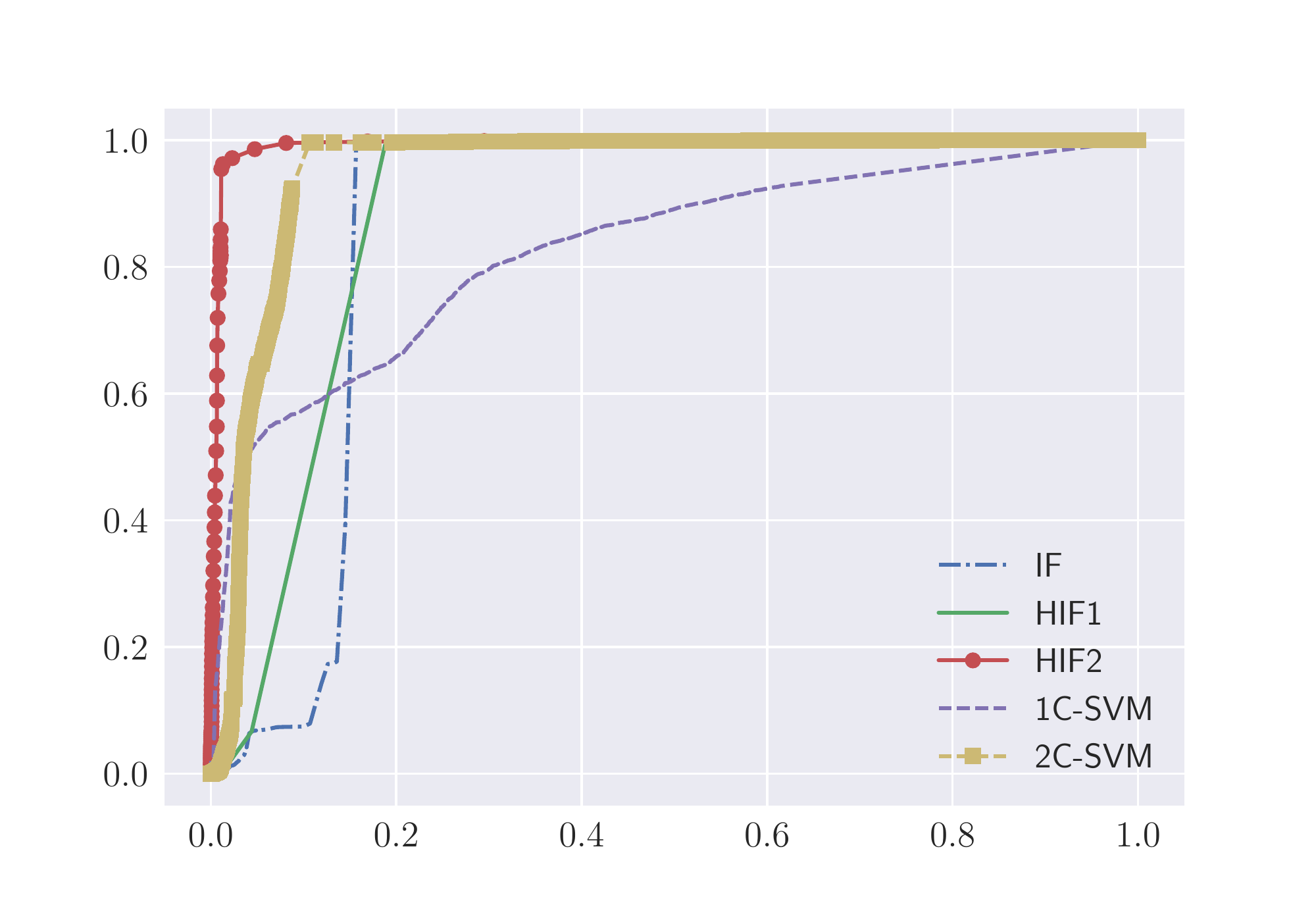}  
\caption{ROC curve obtained on the ISCX HTTPWeb subset for IF (dash-dotted, blue),  HIF1 (solid, green), HIF2 (circle, red), 1C-SVM (dotted, purple), 2C-SVM (dotted, square, yellow)}
\label{fig:iscx_httpweb_roc}
\end{figure}

Figure \ref{fig:iscx_httpweb} presents the distributions of the IF (a) and HIF (b) scores obtained on the ISCX HTTPWeb subset. The learning of 1\% of the labeled anomalies generates a shifting of the distributions that better separates anomalies from 'normal' data. This is also well shown in Figure \ref{fig:iscx_httpweb_roc} which presents the ROC curves for the IF and HIF algorithms.

Furthermore, in Figure \ref{fig:iscx_httpweb}, one can identify a residual anomaly peak that overlaps the 'normal' data distribution. One may try to isolate the data around this peak and ask an experts to labeled some of them to further train the HIF. 

An exploratory data analysis based on the HIF score distributions can thus be used to set up efficiently an interactive semi-supervised procedure, in the scope of an active learning setting.

\begin{figure}[h!]
    \centering
    \includegraphics[width=0.5\textwidth]{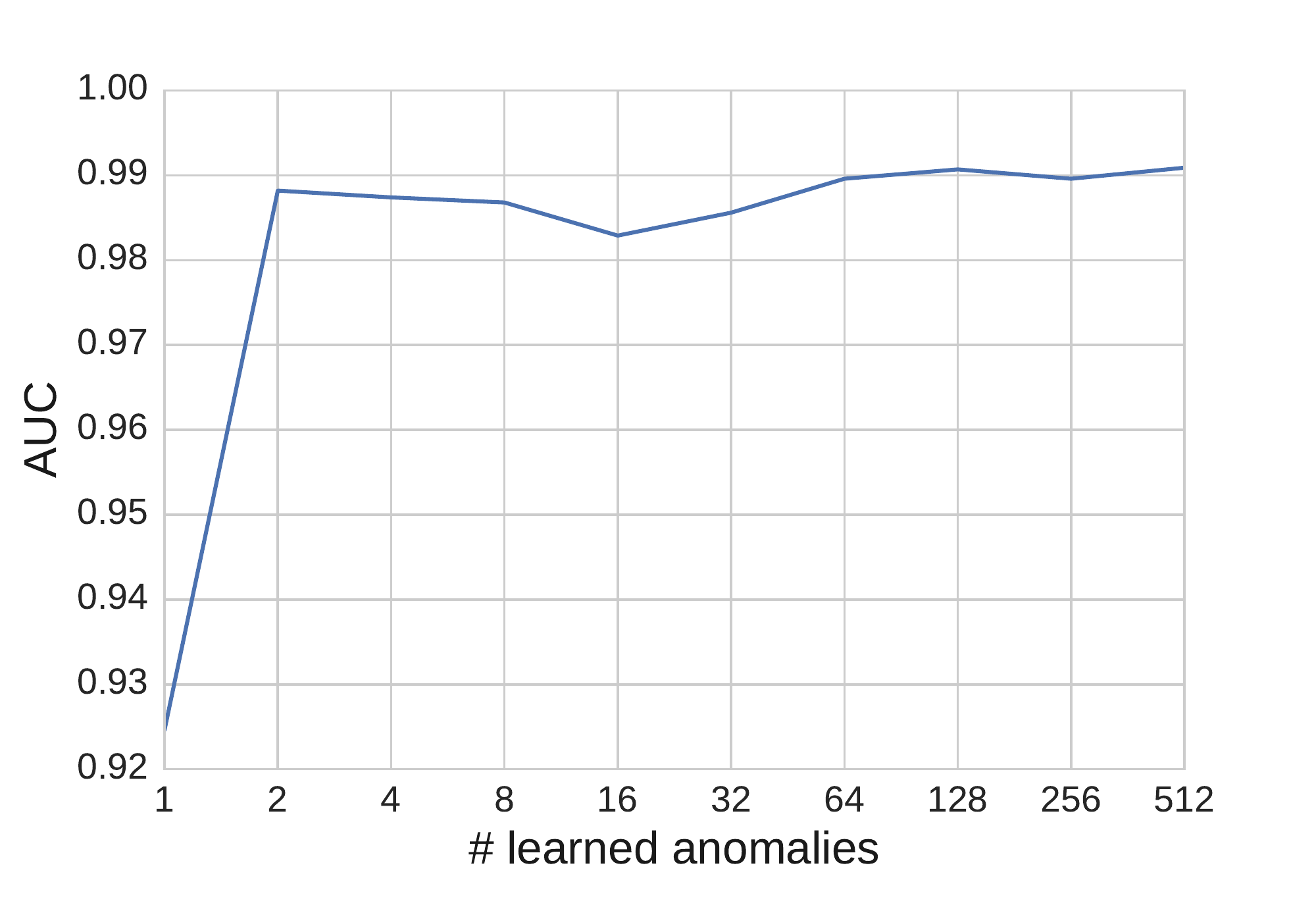} 
 \caption{AUC value as a function of the number of anomalies added into the HIF.  The task consists to separate 'normal' test data from anomalies in the HTTPWeb subset.}
 \label{fig:contaminHTTPWeb}
\end{figure}

Here again, according to Fig.\ref{fig:contaminHTTPWeb}, adding very few labeled anomalies (as low as 2) into the HIF is enough to significantly improve the accuracy of the anomaly detection. This shows the supervised functionality of HIF can be particularly useful when some expertise is available.

\section{Conclusion}

From the construction of a synthetic dataset we have identified an apparently quite penalizing drawback in the Isolation Forest algorithm, the so-called 'blind spots', that characterize unoccupied areas in the data embedding as 'normal' areas even if they are far from the 'normal' data distribution.

To overcome this problem we have proposed a first extension that introduces some centroid based distance calculation in the IF algorithm and shown that the 'blind spot' effect disappears at a very low computational cost.

Furthermore, we have introduced a second extension that provides a supervised capability to the IF, enabling to introduce known anomaly locations into the trees of the isolation forest. A distance based score to these locations is the source to an additional scoring provided by the HIF. A simple linear model has been implemented to aggregate the three anomaly sub-scores proposed in the final Hybrid Isolation Forest (HIF) algorithm.

Extensive testing on a synthetic dataset has been conducted to verify the resolution of the 'blind-spot' effect and to study the impact of the meta parameters used to aggregate the composite scores of HIF, or to estimate the algorithmic complexity of the proposed extensions. 

In addition, the HIF algorithm has been tested on an intrusion detection task. The experimentation carried out on the ISCX benchmark dataset shows  that significant improvements in the detection accuracy can be made by incorporating few known anomalies as train data for the HIF. 

Furthermore, we have shown that the HIF algorithms compare relatively advantageously with the SVM baselines (1C-SVM and 2C-SVM) on the synthetic and ISCX datasets. The apparently successful (although simple) combination of the anomaly detection principle with a supervised classification capability, which is missing in our two baselines, is what makes the HIF original and very competitive at a low complexity cost.  

Finally we believe that the analysis of the distributions of the HIF scores can be used  to set up an interactive semi-supervised or active learning procedure that can help to identify the emergence of new anomaly 'spots' occurring in the data.

\bibliographystyle{plain}
\bibliography{biblio}

\end{document}